\documentclass{article}
\PassOptionsToPackage{numbers, compress}{natbib}
\usepackage[dblblindworkshop,final]{neurips_2025} 
\workshoptitle{AI for Tabular Data}

\usepackage[utf8]{inputenc} 
\usepackage[T1]{fontenc}    
\usepackage{hyperref}       
\usepackage{url}            
\usepackage{booktabs}       
\usepackage{amsfonts}     
\usepackage{nicefrac}       
\usepackage{microtype}      
\usepackage{xcolor}         
\usepackage{amsmath} 
\usepackage{graphicx}
\usepackage{subcaption}
\usepackage{multirow}
\usepackage{placeins}
\usepackage{float}

\title{Causal Data Augmentation for Robust Fine-Tuning of Tabular Foundation Models} 

\author{%
Magnus Bühler \\
Department of Computer Science\\
University of Freiburg\\
Freiburg im Breisgau, Germany\\
\texttt{buehlema@informatik.uni-freiburg.de} \\
\And
Lennart Purucker \\
Department of Computer Science\\
University of Freiburg\\
Freiburg im Breisgau, Germany\\
\And
Frank Hutter \\
Prior Labs\\ 
ELLIS Institute Tübingen\\
University of Freiburg\\
}

\begin{document}
\setlength{\textfloatsep}{6pt plus 2pt minus 2pt}
\setlength{\floatsep}{6pt plus 2pt minus 2pt}
\setlength{\intextsep}{6pt plus 2pt minus 2pt}

\maketitle

\begin{abstract}
Fine-tuning tabular foundation models (TFMs) under data scarcity is challenging, as early stopping on even scarcer validation data often fails to capture true generalization performance. We propose CausalMixFT, a method that enhances fine-tuning robustness and downstream performance by generating structurally consistent synthetic samples using Structural Causal Models (SCMs) fitted on the target dataset. This approach augments limited real data with causally informed synthetic examples, preserving feature dependencies while expanding training diversity. Evaluated across 33 classification datasets from TabArena and over 2{,}300 fine-tuning runs, our CausalMixFT method consistently improves median normalized ROC-AUC by fine-tuning from 0.10 (standard fine-tuning) to 0.12, outperforming purely statistical generators such as CTGAN (-0.01), TabEBM (-0.04), and TableAugment (-0.09). Moreover, it narrows the median validation-test performance correlation gap from 0.67 to 0.30, enabling more reliable validation-based early stopping—a key step toward improving fine-tuning stability under data scarcity. These results demonstrate that incorporating causal structure into data augmentation provides an effective and principled route to fine-tuning tabular foundation models in low-data regimes.
\end{abstract}

\label{chap:abstract}

\section{Introduction}

Foundation models have transformed machine learning across vision~\citep{awais2025foundation_models_in_vision}, medicine~\citep{khan2025foundation_model_in_medicine, azad2023foundation_models_in_medical_imaging}, time series~\citep{kottapalli2025foundation_models_in_time_series}, and graphs~\citep{wang2025foundation_models_for_graphs}. Yet the most ubiquitous data type in the real world, namely \textbf{tabular data}, has long remained the hardest to model effectively. Recent advances in pre-trained tabular foundation models (TFMs) such as TabPFN~\citep{hollmann2022tabpfnv1, Hollmann2025tabpfnv2,garg2025realtabpfn}, TabICL~\citep{qu2025tabicl}, and TabDPT~\citep{ma2024tabdpt} signal a paradigm shift: transformers trained across millions of datasets can now perform in-context learning on unseen tables, rivaling classical methods like XGBoost~\citep{chen2016xgboost}.

While these models demonstrate strong zero-shot generalization, their full potential emerges only after fine-tuning on specific target datasets. Recent models such as Mitra\footnote{\url{https://huggingface.co/autogluon/mitra-classifier}}, TabPFNv2~\citep{Hollmann2025tabpfnv2}, and LimiX~\citep{zhang2025limix} now offer out-of-the-box fine-tuning capabilities in response to the growing demand for data-efficient model adaptation. However, existing fine-tuning practices implicitly assume abundant labeled data, an assumption that is not always met in practice. In fact, more than 39\% of OpenML datasets contain fewer than 1{,}000 samples, representing the largest share of available dataset sizes~\citep{mamdouh2025small_datasets_distribution}. Under such constraints, conventional data splits and validation-based early stopping become unreliable, leading to overfitting to the training or validation split and unstable performance estimates. Consequently, fine-tuning towards small and heterogeneous datasets remains an unsolved challenge for tabular foundation models.

To overcome these limitations, we propose \textbf{CausalMixFT}, a strategy that augments scarce training samples through \emph{learnable Structural Causal Models (SCMs)}. Unlike statistical generators, SCMs recover and exploit causal dependencies among features, producing \textit{structurally consistent} synthetic samples that preserve the semantics of the target domain. By enriching fine-tuning data with causally coherent examples, TFMs can be fine-tuned effectively without overfitting to limited real observations.

\textbf{Contributions.} Our work makes three key contributions:
\begin{enumerate}
    \item \textbf{Empirical diagnosis:} We provide the first systematic analysis of fine-tuning tabular foundation models under severe data scarcity, revealing that large validation-test discrepancies persist even under strong regularization.
    \item \textbf{Methodological innovation:} We introduce \textbf{CausalMixFT}, an SCM-based approach that learns the underlying causal structure of small datasets to generate \textit{structurally consistent} synthetic samples, enabling data-efficient fine-tuning.
    \item \textbf{Comprehensive evaluation:} Across diverse benchmarks and over 2{,}300 fine-tuning runs, our method consistently surpasses conventional fine-tuning and statistical augmentation baselines, establishing a new SOTA for fine-tuning tabular foundation models in low-data regimes.
\end{enumerate}

By combining causal generative modeling with foundation model adaptation, our work provides a principled and data-efficient pathway toward making tabular foundation models fine-tunable in the small-data settings that dominate real-world machine learning.

\label{chap:introduction}

\section{Related Work}
\label{chap:related}
\textbf{PFNs and Variants.} \quad
\citet{muller2021transformers_can_do_bayesian_inferece} introduced \emph{Prior-Data Fitted Networks} (PFNs), using transformers to approximate Bayesian posterior predictive distributions via in-context learning. Subsequent works have extended this paradigm to classification and regression tasks~\citep{hollmann2022tabpfnv1, breejen2024tabforest, ma2024tabdpt, qu2025tabicl, Hollmann2025tabpfnv2, liu2025tabpfnunleashed,zhang2025limix,garg2025realtabpfn}, scaling to larger datasets and diverse pre-training regimes on both real~\citep{ma2024tabdpt} and synthetic data~\citep{breejen2024tabforest}. These advances establish PFNs as universal tabular priors that can generalize across domains, forming the basis for most current TFMs.

\textbf{Fine-Tuning TabPFN \& Regularization.} \quad
Recent work investigates the adaptation of PFNs for large datasets. Approaches include full-weight fine-tuning with prior regularization~\citep{breejen2024tabforest}, continued pre-training~\citep{ma2024tabdpt, garg2025realtabpfn}, tokenization-layer adaptation~\citep{thomas2024retrieval_and_finetuning}, encoder compression and distillation~\citep{feuer2024tunetables, ma2024icl_data_distillation}, mixture-of-experts routing~\citep{shazeer2017mixture_of_experts,xu2024icl_micture_of_experts}, and batch-ensemble encoders~\citep{liu2025tabpfnunleashed}. Additional studies refine context retrieval and conditioning~\citep{thomas2024LoCalPFN, ma2024tabdpt, koshil2024le_tabpfn}, and \citet{rubachev2025on_finetuning_tabFM} provide a general survey. Classical regularization techniques such as L2-SP~\citep{li2018l2sp,garg2025realtabpfn}, stochastic weight averaging~\citep{izmailov2018stocastic_weight_averaging}, and early stopping~\citep{prechelt2002early_stopping_but_when} mitigate overfitting in low-data regimes. However, fine-tuning under \emph{data scarcity}—the regime most common in practice—remains largely unaddressed. While \citet{kadra2021well_tuned_simple_nets} has shown that strong regularization can improve performance in the tabular domain, our work provides a new regularization through data augmentation for TFMs.  

\textbf{Synthetic Tabular Data Generation.} \quad
Synthetic data generation supports privacy, data sharing, and augmentation. GAN-based models~\citep{xu2019ct_gan} and diffusion approaches~\citep{lin2024ctsyn} improve fidelity, while privacy-preserving GANs~\citep{zhao2024ct_gan_plus} and energy-based models~\citep{margeloiu2024tabebm} provide interpretability and control. Yet most focus on distributional realism rather than improving downstream model adaptation. In contrast, we evaluate these approaches for their ability to enhance fine-tuning. 

\textbf{Research Gap.} \quad
Despite progress in tabular foundation models (TFMs) and synthetic data generation, their integration for small-data fine-tuning and regularization remains largely unexplored. Building on \citet{garg2025realtabpfn}, we investigate whether combining real and structured synthetic samples can enhance downstream-performance and robustness. Our work addresses this gap by coupling causal data generation with fine-tuning strategies for TFMs.
\label{chap:related_work}

\section{Methodology}
\label{sec:methodology}

Our method extends the fine-tuning framework of \citet{buhlertowards_finetuning_on_synthetic_data} by mixing real and causally grounded synthetic samples into the fine-tuning process, adding more recent baselines, evaluating a TFM optimized for fine-tuning and evaluating across a broad collection of real world datasets. Specifically, we generate synthetic data using SCMs fitted to the target dataset, enabling the model to learn jointly from real samples and causally coherent augmentations. This design preserves feature dependencies while expanding sample diversity, which enhances robustness and generalization under low-data constraints.

\textbf{SCM-Based Synthetic Augmentation (CausalMixFT).} \quad
Unlike purely statistical generators, SCMs explicitly encode causal dependencies among features through a directed acyclic graph (DAG) and a set of structural equations, allowing data augmentation to respect the underlying data-generating process. We first estimate the structural relations between the features using the PC and FCI algorithms~\citep{spirtes2000PC, spirtes2013FCI}, producing a probabilistic adjacency matrix that encodes edge strengths between variables. DAGs are then sampled and fitted using DoWhy's SCM framework with additive noise models~\citep{dowhy}. Numerical features are modeled with regressors, and categorical features with classifiers. The complexity of the internally used model types can be controlled through a quality hyperparameter. Synthetic samples are generated by sampling exogenous noise and propagating it through the fitted SCM, yielding data that captures both causal structure and realistic variability. See Appendix~\ref{app:generator_details} for more details.

\textbf{Model \& Data Overview.} \quad
We adopt the \emph{Mitra} foundation model as the tabular backbone, as it is explicitly designed for per-dataset fine-tuning and has achieved state-of-the-art performance on the TabArena benchmark. Further, Mitra is provided with strong default hyperparameters. Our experiments cover 33 classification datasets from the \texttt{TabArena} benchmark suite, excluding datasets with more than 200 features (OpenML IDs 46912, 46919, 46939, 46908, 46933) using 10 folds each. Datasets with more than 200 features were excluded to ensure SCM fitting remains within a one-hour runtime limit. Each dataset is split into training, validation, and test subsets using stratified sampling, with training and validation sets capped at 600 and 200 samples respectively to simulate small-data conditions. Further details about the data splitting and light pre-processing are provided in Appendix~\ref{sec:data_splitting} and \ref{sec:data_preprocessing}.

\textbf{Fine-Tuning and Implementation.} \quad
\begin{figure}[t]
  \centering
  \includegraphics[width=0.96\textwidth]{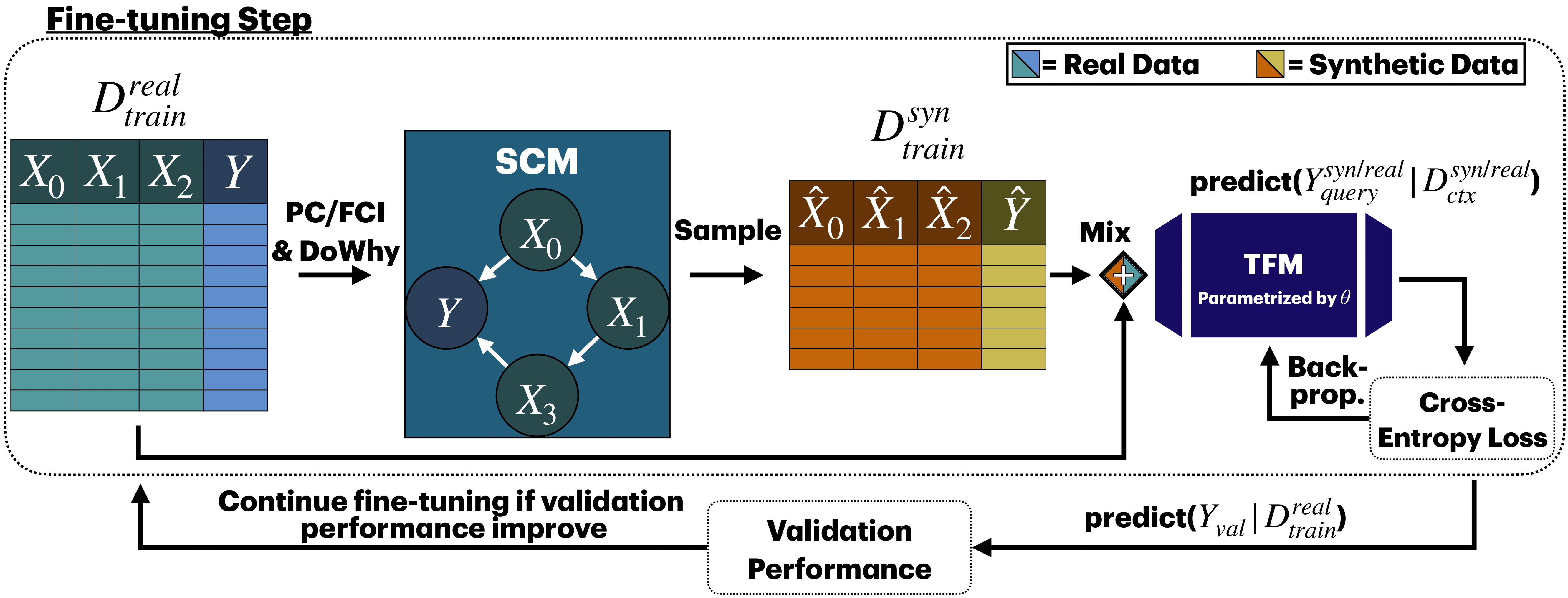}
  \caption{
  \textbf{Overview of the SCM-augmented fine-tuning process.}
  Real training data ($D^{\text{real}}_{\text{train}}$) are used to fit a Structural Causal Model (SCM) via PC/FCI and DoWhy~\citep{dowhy}.
  The SCM samples synthetic data ($D^{\text{syn}}_{\text{train}}$) that preserve the discovered causal dependencies among features.
  Real and synthetic samples are mixed in equal proportion to fine-tune the tabular foundation model (TFM), which is optimized by cross-entropy loss.
  Validation is performed only on real data, and fine-tuning continues as long as validation performance improves.
  }
  \label{fig:scm_pipeline}
\end{figure}
Let the downstream dataset be denoted as $D^{\text{real}} = \{(x_i, y_i)\}_{i=1}^{n}$ and the corresponding SCM-generated data as $D^{\text{syn}} = \{(x_j^{\text{syn}}, y_j^{\text{syn}})\}_{j=1}^{m}$. The model is fine-tuned on the combined dataset $D^{\text{mix}} = D^{\text{real}} \cup D^{\text{syn}}$, where both sources are equally represented in each batch. To balance contributions from real and synthetic data, we define a weighted fine-tuning objective:
\[
\mathcal{L} = \alpha \, \mathbb{E}_{(x, y) \sim D^{\text{real}}}[\ell(f_{\theta'}(x), y)] + (1 - \alpha) \, \mathbb{E}_{(x, y) \sim D^{\text{syn}}}[\ell(f_{\theta'}(x), y)],
\]
where $\alpha = 0.5$ unless specified otherwise. Validation is performed exclusively on $D^{\text{real}}_{\text{val}}$ to ensure that improvements reflect genuine generalization rather than memorization of synthetic data. Early stopping is triggered when validation log-loss fails to improve for a fixed number of iterations. Optimization uses the default \emph{Mitra} hyperparameters. A schematic overview of the iterative fine-tuning steps is shown in Figure~\ref{fig:scm_pipeline}. Additional implementation details and hyperparameter settings are provided in Appendix~\ref{app:generator_details}.

\label{chap:methodology}

\section{Results}
\label{sec:results}

We evaluate whether CausalMixFT improves the robustness and generalization of tabular foundation models under data scarcity. Experiments are conducted on the \emph{Mitra} model across 33 classification datasets with 10 folds each from the \texttt{TabArena} benchmark suite, totaling 2,310 fine-tuning runs. Model performance is reported as normalized ROC-AUC relative to the pre-trained model (see Appendix~\ref{para:perfomance_normalization}).

\begin{figure}[t]
  \centering
  \begin{subfigure}[t]{0.48\textwidth}
    \centering
    \includegraphics[width=\textwidth]{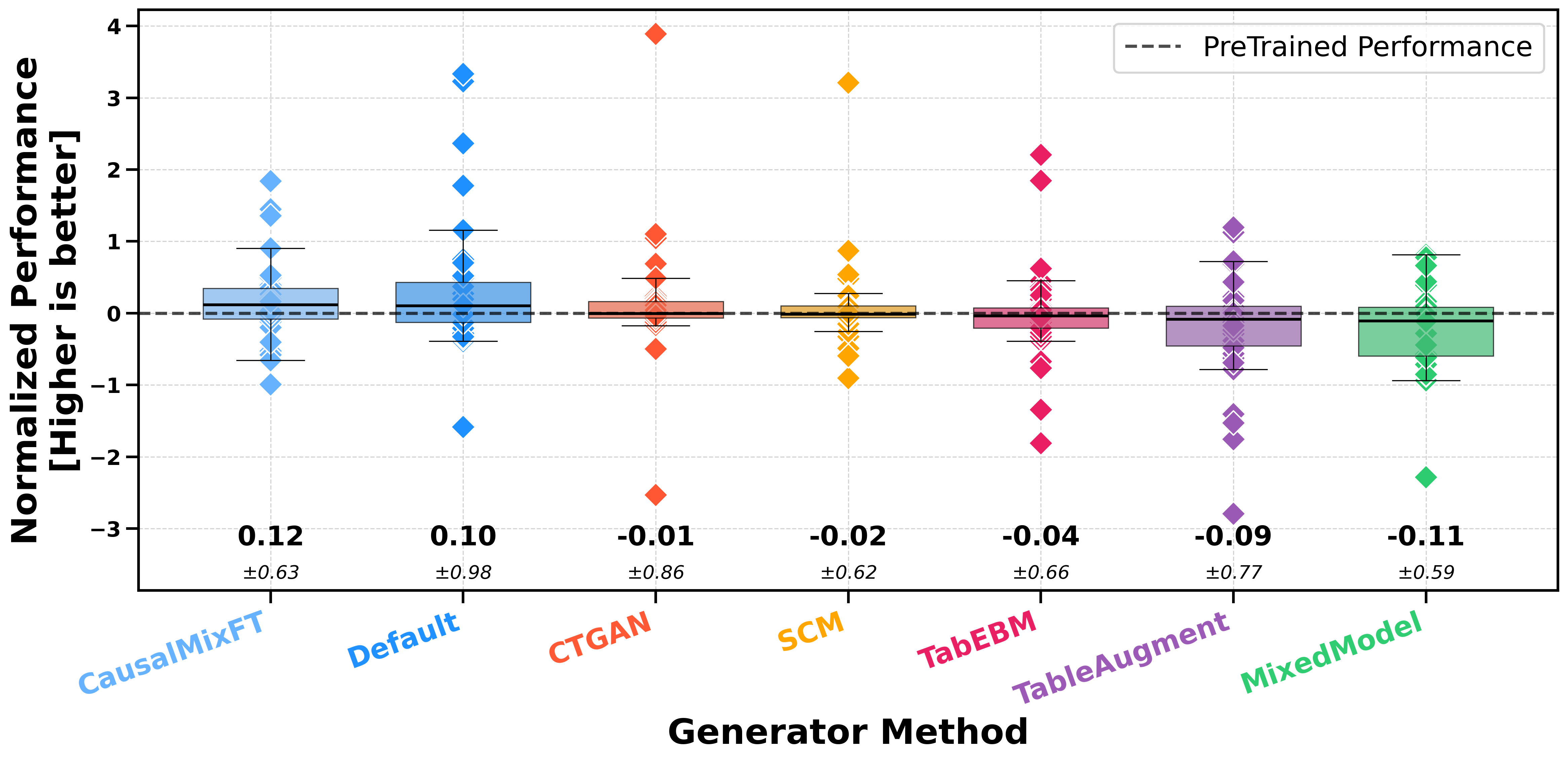}
  \end{subfigure}
  \hfill
  \begin{subfigure}[t]{0.48\textwidth}
    \centering
    \includegraphics[width=\textwidth]{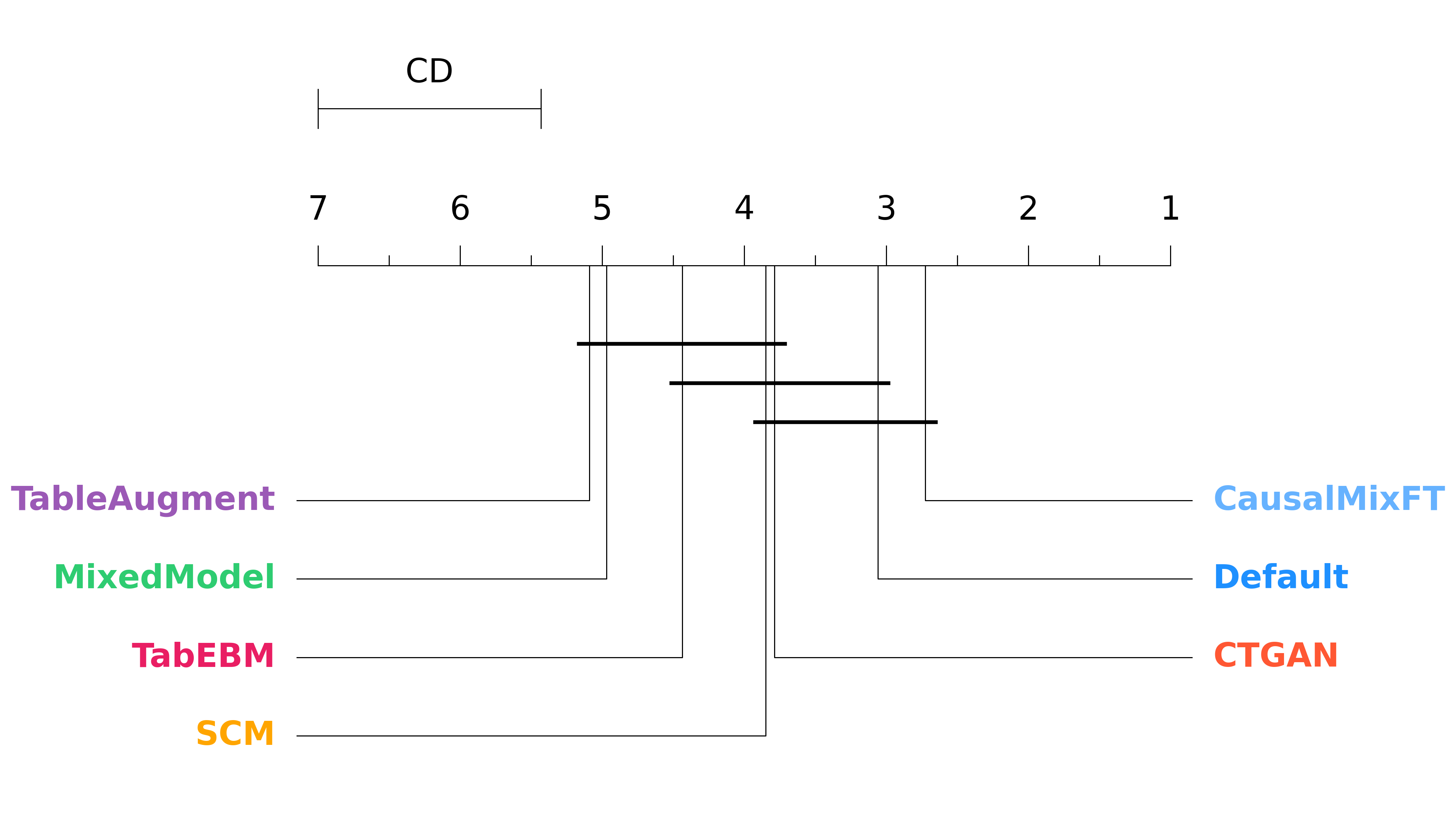}
  \end{subfigure}
  \caption{
  \textbf{Performance comparison across data generation strategies.}
  (left) Normalized ROC-AUC improvements relative to the pre-trained baseline (dashed line).
    Whiskers represent $1.5\times$ IQR; medians and standard deviations are annotated. \textbf{Higher score is better}. 
  (right) Critical difference diagram (significance level = 0.05) \textbf{Lower rank is better}~\citep{demvsar2006ranking_method}.
  }
  \label{fig:results_combined}
\end{figure}

\textbf{Fine-Tuning Performance.}\quad Figure~\ref{fig:results_combined} summarizes the normalized test performance across all data generation strategies. On the left plot the proposed CausalMixFT, which combines real and causally generated samples, achieves the highest median improvement of (\textbf{+0.12}±0.63) over the pre-trained model, outperforming both the default fine-tuning baseline (+0.10 ±0.98) and all purely synthetic augmentation methods, including CTGAN, SCM, TabEBM, TableAugment and MixedModel (which, in fact, show negative median improvements).

While default fine-tuning occasionally achieves higher peak performance on individual datasets, its variability is substantially larger than that of our method \textbf{(Default: ±0.98 vs. CausalMixFT: ±0.63)}. This indicates greater instability across datasets, whereas the \emph{CausalMixFT} configuration acts as a consistent regularizer, improving median performance and reducing variance. These results show that SCM-based augmentation stabilizes fine-tuning under small-data conditions by introducing causally structured synthetic diversity.

Figure~\ref{fig:results_combined} (right) presents the average ranks and corresponding critical difference (CD) intervals across datasets. \textbf{CausalMixFT ranks first overall}, followed by the default fine-tuning baseline, while purely synthetic generators occupy lower ranks, confirming the results of the boxplot. We further analyze the validation-test performance gap in Appendix~\ref{sec:val_test_performance_gap}, showing that early stopping based on limited validation data leads to significant validation set overfitting depending on the fine-tuning data mix used.

\section{Discussion}
We empirically find that fine-tuning tabular foundation models (TFMs) in tiny-to-small data regimes remains highly challenging. To address this, we propose \textbf{CausalMixFT}, a causally grounded fine-tuning approach that leverages SCM-based augmentation to improve both stability and generalization. Across a curated benchmark, \textbf{CausalMixFT} consistently outperforms standard fine-tuning, achieving a superior balance between robustness and data efficiency. This provides a principled path forward for more reliable adaptation of TFMs under scarce supervision.

\label{chap:results}

\clearpage

\bibliographystyle{plainnat}
\bibliography{bib}          

\textbf{Acknowledgment}
The authors acknowledge the use of ChatGPT-5 (OpenAI, 2025) for assistance in refining sentence formulations and in structuring tables and figures to enhance the clarity and presentation of this paper.
\newpage
\appendix
\section*{Appendix Overview}
This appendix provides additional analyses, figures, and implementation details that complement the main text. 
It includes extended evaluations of validation-test correlations, overfitting behavior, performance heterogeneity across data generators, as well as in-depth analyses of model weight adaptation and normalization procedures.

\section{Validation--Test Performance Gap}
\label{sec:val_test_performance_gap}

To better understand the relationship between validation and test performance during fine-tuning, we analyze the Pearson correlation between validation log-loss and test log-loss across all generator configurations. While validation metrics often suggest strong improvements, the corresponding test results frequently show diminished gains. This discrepancy indicates that validation performance provides a weak and noisy signal for true generalization, particularly in the low-data regime where validation splits are small.

\begin{figure}[t]
  \centering
  \includegraphics[width=0.96\textwidth]{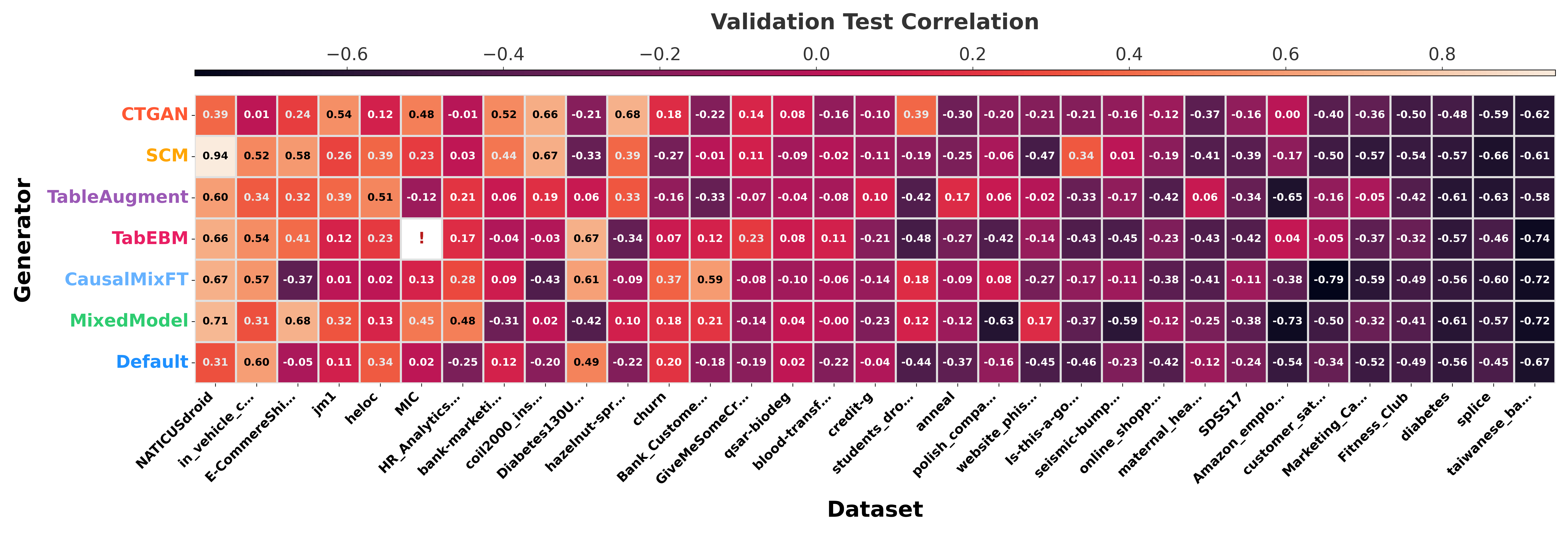}
  \caption{\textbf{Validation--test performance correlation across datasets and data generators.}
  Each cell shows the Pearson correlation between validation and test log-loss for a given dataset and generator configuration. 
  Low or negative correlations indicate that validation performance is not a reliable proxy for generalization under small-data conditions. The columns and rows are sorted by average correlation coefficients from left (higher) to right (lower) and top (higher) to bottom (lower). Incomplete runs due to the time limit of 1h or too many features are marked with "\textcolor{red}{!}" (TabEBM uses TabPFNv1 internally, which only allows for 100 features)}
  \label{fig:val_test_correlation}
\end{figure}

Figure~\ref{fig:val_test_correlation} presents the correlation heatmap between validation and test performance across datasets and generator types. Correlations vary substantially, with several negative or near-zero values, highlighting the instability of validation-based early stopping under data-scarce conditions. Among all methods, the \textit{CausalMixFT} configuration yields relatively higher and more consistent correlations, suggesting that incorporating causally structured synthetic data mitigates some of this instability. Nonetheless, across most settings, validation performance remains an unreliable predictor of test performance, underscoring the need for more robust fine-tuning criteria for tabular foundation models. The general low generalization from validation to test performance is one of the main factors we identified for fine-tuning TFMs to generally be very challenging. 

\section{Validation Performance and Overfitting Analysis}
\label{sec:validation_performance}

To complement the test-set evaluation, we analyze the normalized ROC-AUC performance on the validation sets across all generator configurations. Comparing validation and test performance provides insight into the degree of overfitting introduced during fine-tuning and the reliability of validation metrics as an early-stopping signal. A smaller discrepancy between validation and test performance indicates a more stable and trustworthy proxy for generalization.

\begin{figure}[t]
  \centering
  \begin{subfigure}[t]{0.48\textwidth}
    \centering
    \includegraphics[width=\textwidth]{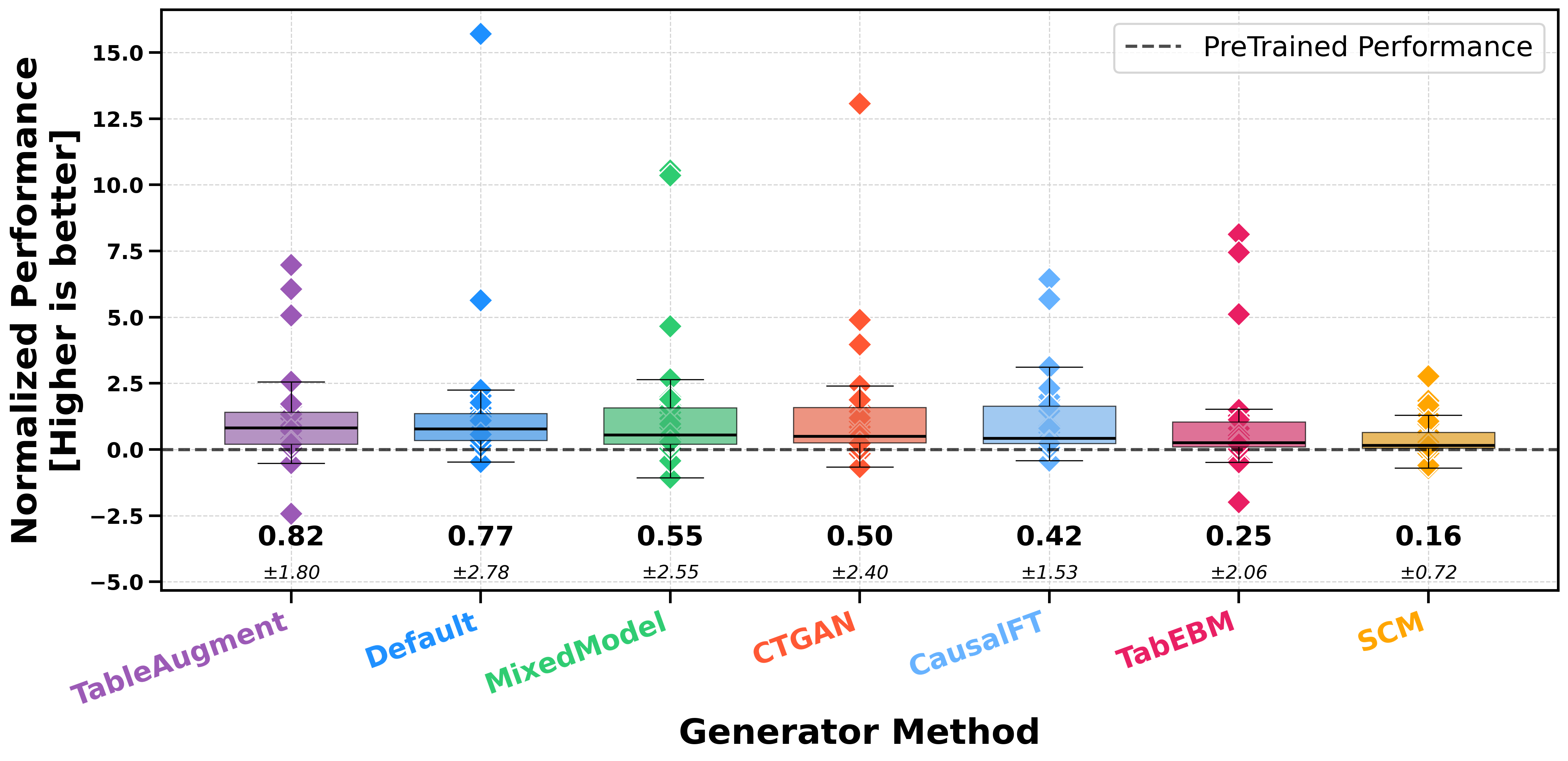}
    \caption{\textbf{Validation ROC-AUC distribution across generator methods.} Each box represents the normalized validation performance relative to the pre-trained baseline.}
    \label{fig:validation_boxplot_results}
  \end{subfigure}
  \hfill
  \begin{subfigure}[t]{0.48\textwidth}
    \centering
    \includegraphics[width=\textwidth]{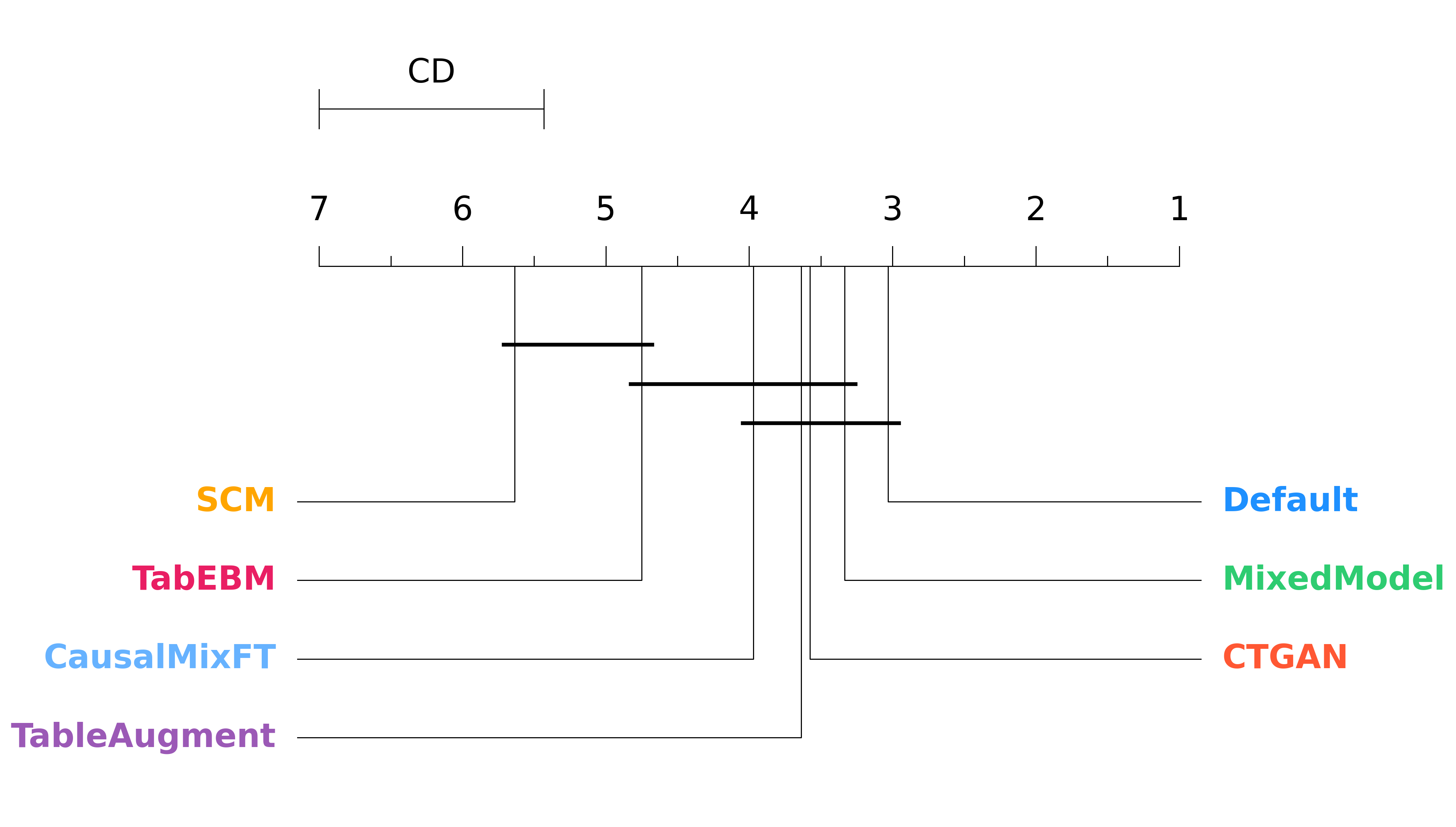}
    \caption{\textbf{Critical difference diagram (significance level = 0.05).} Lower ranks indicate better average performance across datasets.}
    \label{fig:validation_autorank}
  \end{subfigure}
  \caption{\textbf{Validation performance comparison across generators.} 
  The results reveal that validation-based ranking can be misleading under small-data conditions, with methods such as \textit{TableAugment} and \textit{Default} showing strong validation performance but large decrease in test performance and thus generalization. 
  The \textit{CausalMixFT} configuration demonstrates a smaller validation-test discrepancy, suggesting more stable and generalizable fine-tuning behavior.}
  \label{fig:validation_results}
\end{figure}

Figure~\ref{fig:validation_results} summarizes validation performance and the corresponding ranks across generators. Figure~\ref{fig:validation_boxplot_results} reveals that the \textit{TableAugment} generator achieves the highest median normalized validation ROC-AUC, despite ranking among the weakest methods on the test set (Section~\ref{sec:heterogeneity_across_generators}). This discrepancy suggests strong overfitting to the validation data, leading to poor generalization. Similarly, the default fine-tuning baseline configuration exhibits the second-highest validation performance but also a pronounced drop on the test set, with a median validation-test difference of 0.67 normalized units. In contrast, the \textit{CausalMixFT} combination shows a much smaller difference of 0.30 units, indicating that SCM-based augmentation produces a more reliable and stable validation signal.

The critical difference diagram in Figure~\ref{fig:validation_autorank} further supports these observations. The \textit{Default} baseline achieves the lowest (best) rank on validation performance, reflecting its overconfident behavior on small validation splits. However, this high validation ranking does not translate into superior test performance, reinforcing the conclusion that validation metrics can be misleading under data-scarce conditions. Across all methods, the majority of validation ROC-AUC scores exceed those of the pre-trained model, which to an extent is an expected outcome due to early stopping. Certain datasets display lower validation performance, relative to the pre-trained model, likely caused by divergence between the early-stopping criterion (log-loss) and the evaluation metric (ROC-AUC).

\section{Heterogeneity of Test Performance across Generators}
\label{sec:heterogeneity_across_generators}

To investigate how fine-tuning outcomes differ across data generation strategies, we analyze the normalized test ROC-AUC for each dataset and generator combination. This evaluation highlights the degree of heterogeneity in model performance and the dataset-specific behavior of each augmentation method. We note that the TabEBM generator is not compatible with the "MIC" dataset, as it internally relies on TabPFNv1, which only supports up to 100 features, while the dataset has 112 features. 

\begin{figure}[t]
  \centering
  \includegraphics[width=0.96\textwidth]{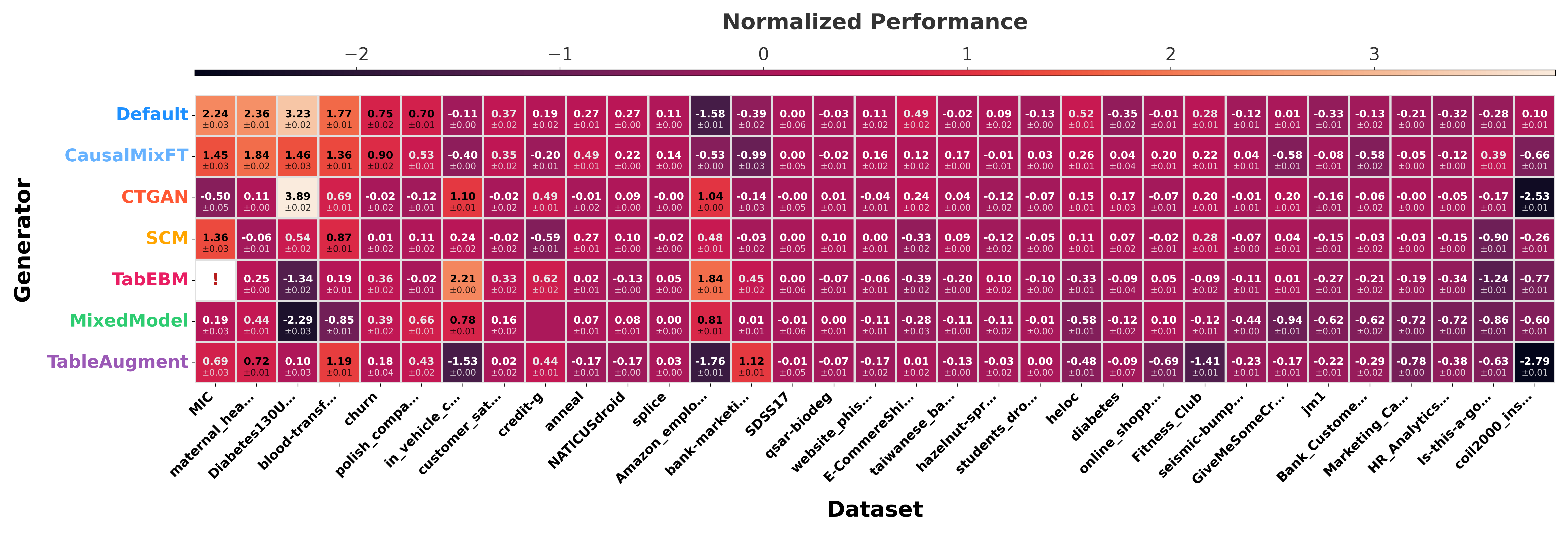}
  \caption{\textbf{Normalized test ROC-AUC performance across datasets and generator configurations.}
  Each cell reports the normalized ROC-AUC (mean ± standard deviation) for a given generator on a specific dataset. 
  The observed heterogeneity indicates that fine-tuning performance varies considerably across generators and datasets, 
  highlighting the need for dataset-adaptive augmentation strategies.}
  \label{fig:test_rocauc_finetunability}
\end{figure}

Figure~\ref{fig:test_rocauc_finetunability} presents the normalized test performance heatmap across all generators and datasets. Consistent with the findings in Section~\ref{sec:results}, we observe substantial variability in fine-tuning outcomes. While the \textit{CausalMixFT} and \textit{Default} configurations achieve strong and stable performance on average, their relative advantage varies across datasets. Some datasets favor purely synthetic approaches such as \textit{CTGAN} or \textit{TabEBM}, whereas others benefit most from hybrid or causally informed augmentation. The \textit{CTGAN} generator achieves highes normalized performance of 3.89, while the TableAugment generator yields the lowest normalized performance of -2.79. 

This heterogeneity suggests that the effectiveness of a generator is strongly dataset dependent and influenced by the underlying feature distribution, sample size, and causal complexity. The absence of a universally superior generator underscores the importance of adaptive fine-tuning strategies that can leverage multiple synthetic sources or dynamically adjust augmentation ratios based on dataset characteristics.

\section{Combinations of Generators}
\label{sec:generator_combinations}

Building on the strong performance of our proposed SCM-based augmentation, we extend our analysis to explore hybrid generator configurations that combine multiple data sources. Given the competitive baseline performance of the default fine-tuning setup (\textit{Default}) relative to single synthetic generators, we systematically pair it with one or more synthetic generation methods to examine potential complementarity effects. We evaluate normalized test ROC-AUC performance across all datasets to assess whether combining generators yields more robust fine-tuning behavior.

\begin{figure}[t]
  \centering
  \includegraphics[width=0.96\textwidth]{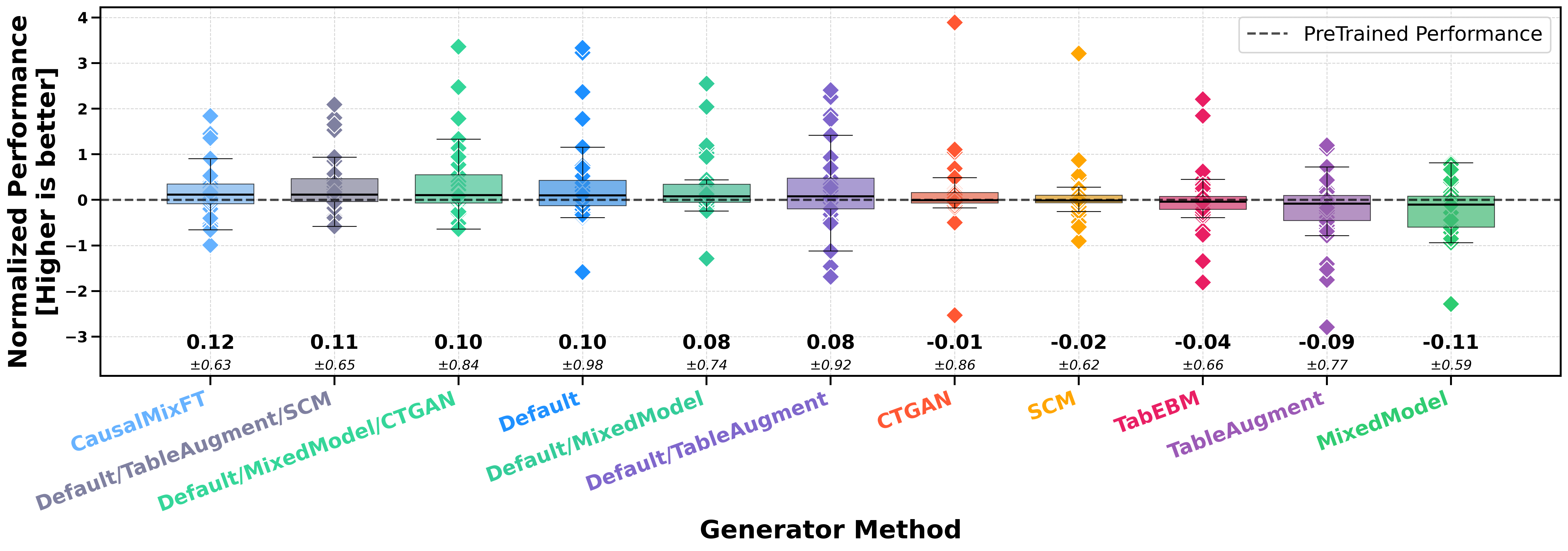}
  \caption{\textbf{Normalized test ROC-AUC performance across mixed generator configurations.}
  Each box represents the distribution of normalized test performance relative to the pre-trained model across datasets.
  Mixed configurations that combine real and synthetic data consistently outperform single-generator baselines.}
  \label{fig:all_generators_test_rocauc_boxplot}
\end{figure}

As shown in Figure~\ref{fig:all_generators_test_rocauc_boxplot}, hybrid configurations that mix real and synthetic data achieve consistently strong results across datasets. The five mixed-generator variants occupy the top six median performance positions, with \textit{CausalMixFT} achieving the highest median improvement of \textbf{+0.12} (±0.63), followed closely by \textit{Default/TableAugment/SCM} (\textbf{+0.11} ±0.65). The latter finding is particularly noteworthy, as \textit{TableAugment} alone performs poorly when used in isolation. This suggests that combining heterogeneous data sources enables the foundation model to leverage complementary structural and distributional properties—an effect reminiscent of ensemble learning.

Furthermore, the \textit{Default/MM} configuration exhibits both competitive median performance (\textbf{+0.10}) and a notably small interquartile range (1.5 $\times$IQR) shown by the whiskers, indicating stable and predictable improvements across datasets. Such stability makes it a strong candidate when reliable gains over the pre-trained baseline are prioritized alongside generalization consistency. Overall, these results demonstrate that mixing real data with diverse synthetic generators enhances fine-tuning robustness and generalization, highlighting the potential of multi-generator augmentation for tabular foundation models operating in the low-data regime.

\section{Fine-Tuned Weight Distance from the Pre-Trained Model}
\label{sec:weight_distance}

We analyze how far the fine-tuned model weights deviate from the pre-trained checkpoint across different generator configurations. Specifically, we compute the elementwise Euclidean distance between the fine-tuned parameters and the original pre-trained weights. Prior work has suggested that constraining the degree of weight divergence through regularization (e.g. euclidean distance) can improve fine-tuning stability and generalization~\citep{garg2025realtabpfn, li2018l2sp}. By quantifying the total weight displacement per generator, we aim to understand how different augmentation strategies influence model adaptation dynamics and parameter stability.

\begin{figure}[t]
  \centering
  \includegraphics[width=0.96\textwidth]{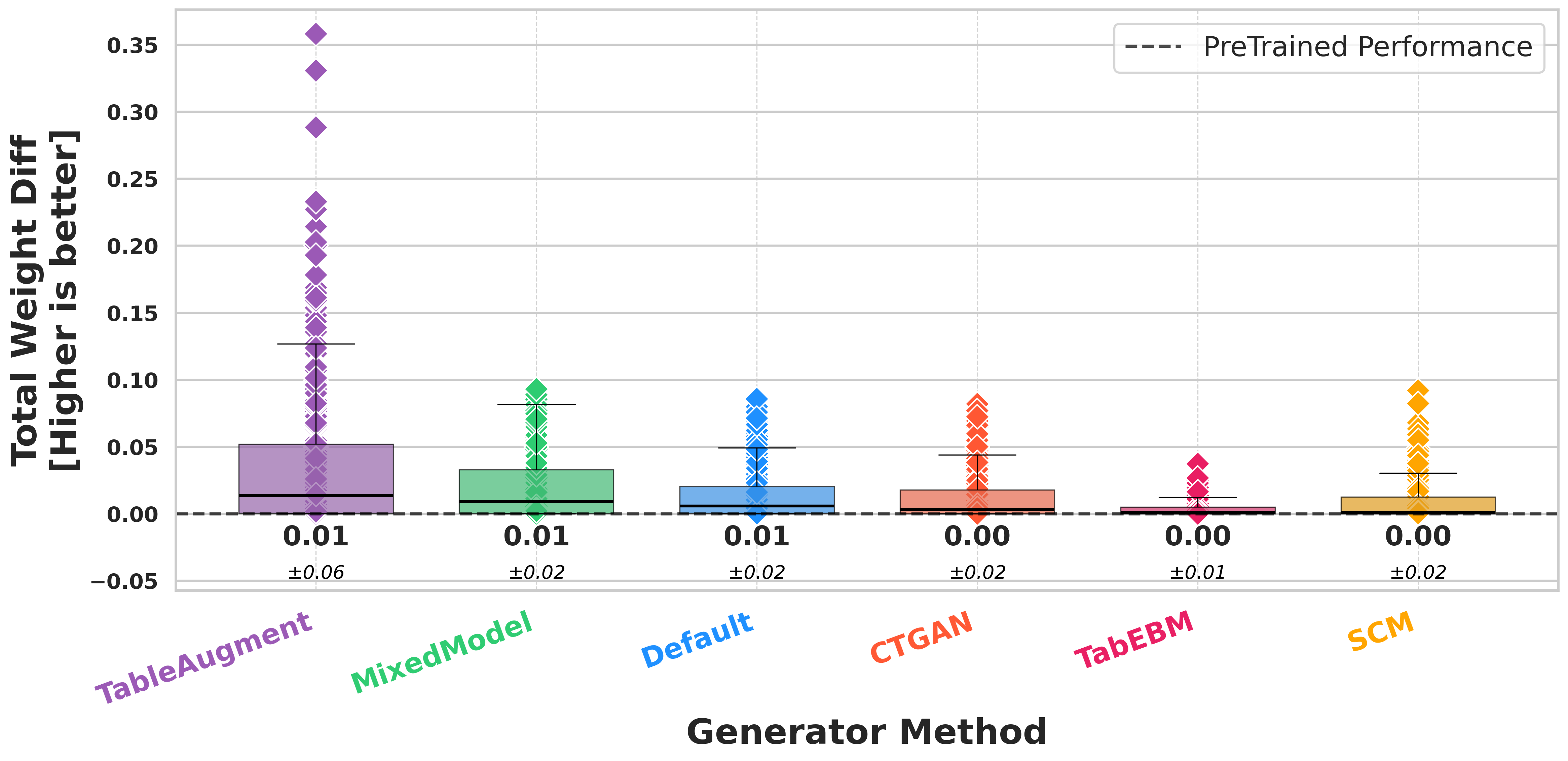}
  \caption{\textbf{Total parameter distance between fine-tuned and pre-trained model weights across generator configurations.}
  Each box represents the distribution of elementwise Euclidean weight differences across all datasets. 
  Smaller values indicate that the fine-tuned model remains closer to the pre-trained parameter space.}
  \label{fig:average_weight_distance_per_generator}
\end{figure}

As shown in Figure~\ref{fig:average_weight_distance_per_generator}, the \textit{TableAugment} generator exhibits the largest variance in weight displacement from the pre-trained model. Although its median distance is relatively low (approximately $0.01$), several runs show extreme deviations up to $0.35$, indicating unstable optimization behavior. Combined with its previously observed weak generalization performance, this pattern is consistent with phenomena associated with \emph{catastrophic forgetting}~\citep{kotha2023understandingcatastrophicforgetting, bethune2025scaling_laws_catastrophic_forgetting, goodfellow2013empirical_catastrophic_forgetting}, where fine-tuning overwrites pre-trained representations, leading to significant loss of learned capabilities.

In contrast, the \textit{MixedModel}, \textit{Default}, and \textit{CTGAN} configurations display similar median distances (around $0.01$) but diverge substantially in their downstream performance, suggesting that weight distance alone is an unreliable indicator of fine-tuning success. Notably, the \textit{TabEBM} and \textit{SCM} generators show almost no displacement from the pre-trained weights, implying that their training signals were too weak or misaligned to induce meaningful parameter updates. This observation highlights that small weight changes do not necessarily imply better generalization, emphasizing the importance of evaluating both representational stability and downstream performance jointly.

\section{Layer wise Weight Adaptation}
\label{sec:layerwise_weight_adaptation}

We next examine which components of the TFM undergo the largest updates relative to the pre-trained checkpoint. 
The pre-trained model has been optimized on a broad distribution of purely synthetic tasks, which encourages storage of general representational structure rather than dataset-specific heuristics. 
Fine-tuning, in contrast, repeatedly exposes the model to related (or even equal) samples from a generator, which promotes specialization towards these samples. 
Quantifying where parameter drift concentrates within the network can therefore provide insight into how specialization emerges.

\begin{figure}[t]
  \centering
  \includegraphics[width=0.96\textwidth]{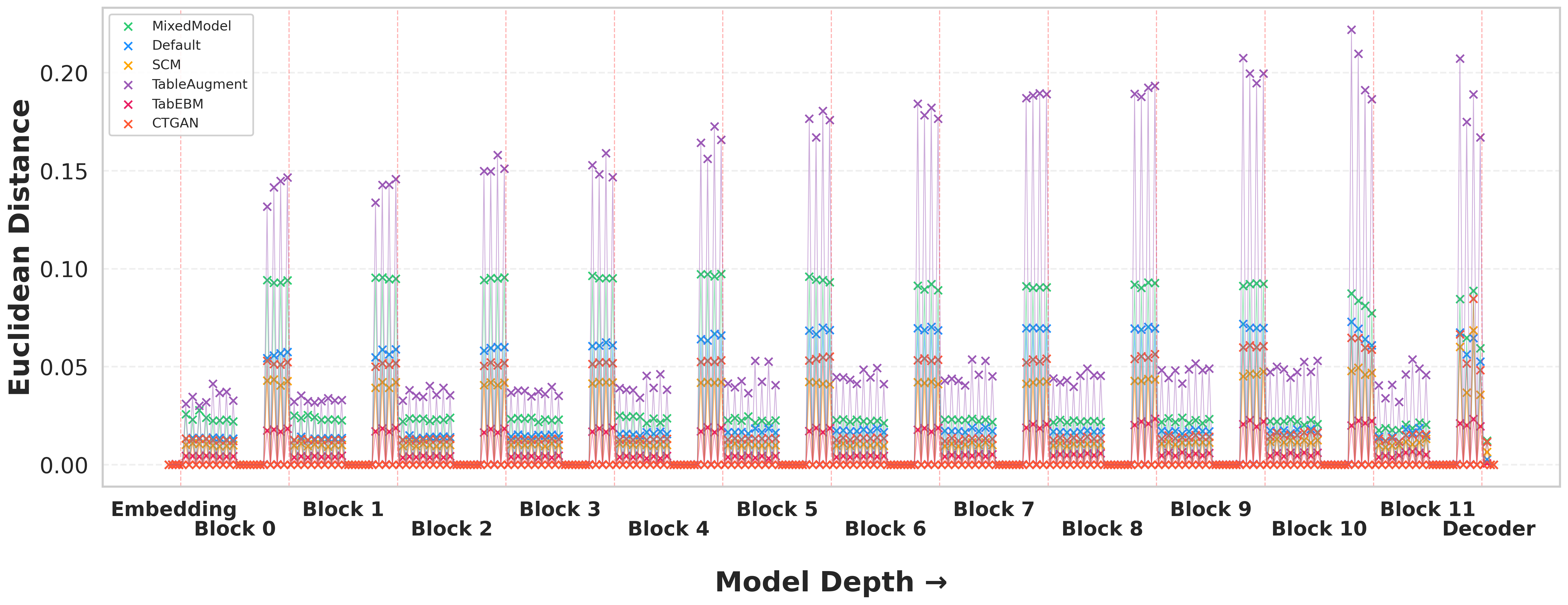}
  \caption{\textbf{Layerwise parameter drift by generator.}
  Euclidean distance between fine-tuned and pre-trained parameters, grouped by module along model depth from left to right 
  (embedding, transformer blocks $0$ through $11$, decoder head). 
  Markers correspond to distinct parameter groups within each block.}
  \label{fig:weight_distance_per_generator_across_model}
\end{figure}

Figure~\ref{fig:weight_distance_per_generator_across_model} shows the distribution of layerwise weight distances for each generator. 
The \textit{TableAugment} configuration displays a pronounced depth trend: 
distances increase toward later blocks and the decoder head, indicating strong adaptation in task-specific layers. 
A similar pattern, though less pronounced, appears for the \textit{Default}, \textit{SCM}, and \textit{CTGAN} settings. 
In contrast, \textit{MixedModel} and \textit{TabEBM} show relatively flat profiles with smaller shifts across depth. 
These observations align with the common view that early layers encode general computations while deeper layers encode task-specific transformations; 
specialization during fine-tuning therefore concentrates in later blocks.

Coupling these results with Section~\ref{sec:validation_performance} suggests a practical implication. 
When depth-wise drift is steep and accompanied by weak test generalization, as observed for \textit{TableAugment}, 
regularization that limits deviation from the pre-trained manifold may improve stability of early stopping and reduce overfitting.

\section{Component-Wise Weight Adaptation}
\label{sec:componentwise_adaptation}

The layer wise analysis in Figure~\ref{fig:weight_distance_per_generator_across_model} indicates that certain model components undergo substantial parameter updates during fine-tuning, whereas others remain largely unchanged across generators. To investigate this in greater detail, we isolate and compare the two component groups exhibiting the strongest deviations from the pre-trained checkpoint.

\begin{figure}[t]
  \centering
  \begin{subfigure}[t]{0.48\textwidth}
    \centering
    \includegraphics[width=\textwidth]{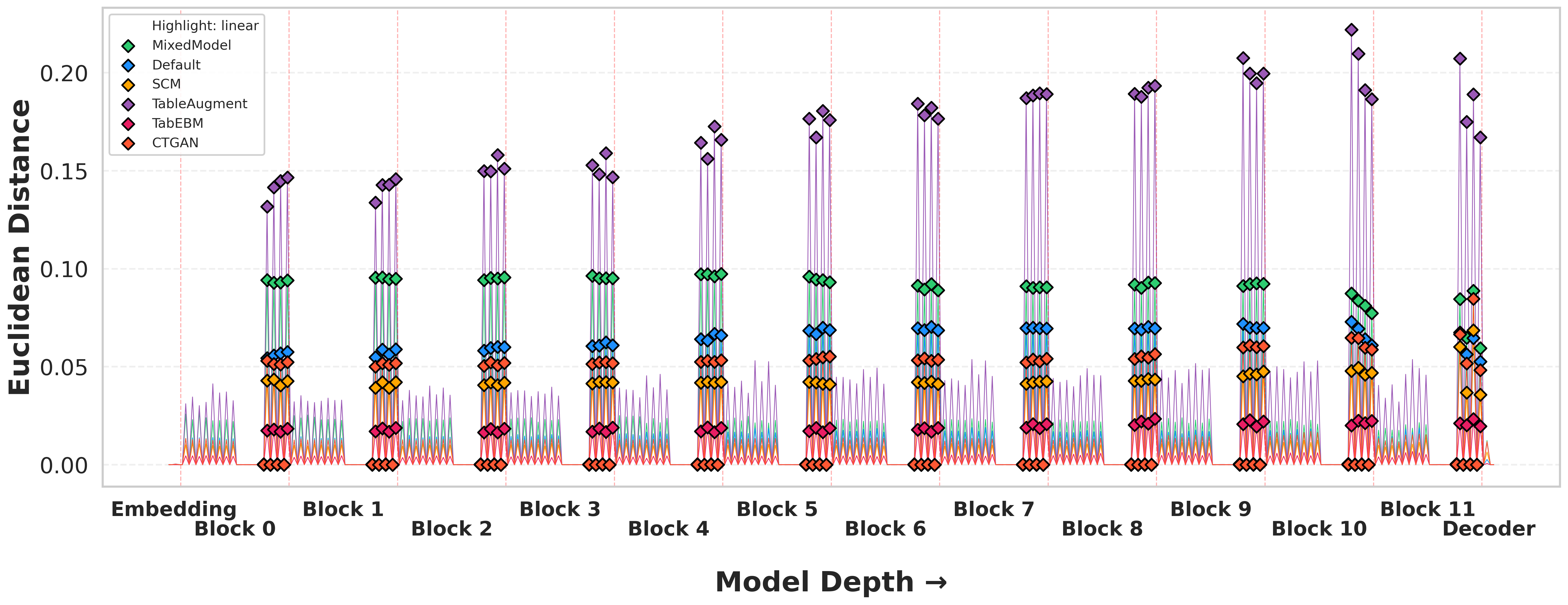}
    \caption{\textbf{Feed-forward (linear) layers.} Weight distance across model depth per generator.}
    \label{fig:linear_weight_distance_per_generator_across_model}
  \end{subfigure}
  \hfill
  \begin{subfigure}[t]{0.48\textwidth}
    \centering
    \includegraphics[width=\textwidth]{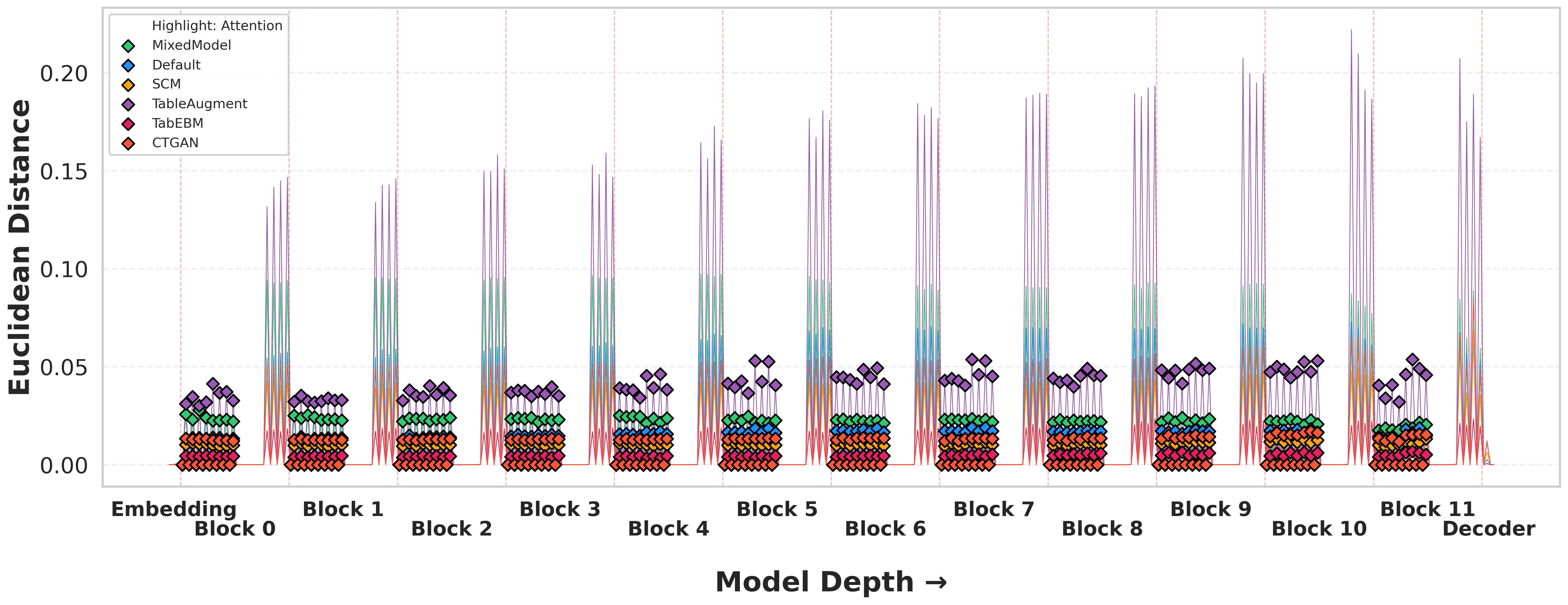}
    \caption{\textbf{Attention layers.} Weight distance across model depth per generator.}
    \label{fig:attention_weight_distance_per_generator_across_model}
  \end{subfigure}
  \caption{\textbf{Component-wise weight deviation across model depth.} 
  Comparison of parameter shifts for feed-forward (linear) and attention layers relative to the pre-trained model. 
  Linear layers show the strongest adaptation across all generators, followed by the attention layers.}
  \label{fig:linear_and_attention_weight_distance_per_generator_across_model}
\end{figure}

Figure~\ref{fig:linear_and_attention_weight_distance_per_generator_across_model} highlights that the \emph{linear layers}, including the feed-forward networks between each transformer block. We observe that especially the feed-forward networks between the attention computation undergo the largest parameter shifts across all generator configurations, while the decoder head has (not marked on the plot) only minimal distance compared to the pre-trained model weights. This suggests that most dataset-specific specialization is concentrated in the linear layers, while other components retain more general representations. These findings point to a potential strategy for mitigating overfitting: selectively constraining the deviation of linear layer parameters from the pre-trained checkpoint may preserve adaptability in other components while reducing excessive specialization. Such a targeted regularization offers an interesting direction for future research.

The \emph{attention layers} (key, query, value, and output projection matrices) exhibit the second-largest parameter shifts, consistent with earlier findings that fine-tuning induces measurable but in our case less pronounced adaptation in attention modules~\citep{rubachev2025on_finetuning_tabFM}. This supports the hypothesis that attention mechanisms encode more generalizable computations that remain relatively stable across datasets. Notably, the extent of attention-layer adaptation varies strongly with the generator type: \textit{TableAugment} produces the largest deviations, followed by the \textit{MixedModel}, while \textit{Default}, \textit{CTGAN}, \textit{TabEBM}, and \textit{SCM} lead to comparatively minor changes. These results further reinforce that the fine-tuning signal introduced by different generators differentially influences the degree to which the model parameters are adjusted.

\section{Performance Normalization}
\label{para:perfomance_normalization}
to compare the performance across different data generators, we apply the normalization strategy suggested by \citet{gorishniy2024tabm}. We choose the base model's (Mitra's) zero-shot performance as performance baseline, to measure the improvement after fine-tuning over the pre-trained model. To normalize the performance of the fine-tuned model we compute: $score_{normalized} = metric_{sign} \times(\frac{score_{method}}{score_{baseline}} - 1) \times 100\%$, where $metric_{sign} = 1$ for metrics, where higher is better (e.g. ROC-AUC) and $metric_{sign} = -1$ for metrics, where lower is better (e.g. Log-loss). If fine-tuning and the pre-trained models achieve the same performance then $score_{normalized} = 0$, if fine-tuning improves over the pre-trained model then $score_{normalized} > 0$ and if fine-tuning decreases performance then $score_{normalized} < 0$. The choice of normalization method allows averaging the normalized performance across datasets and compare the data generating methods.

\section{Notation.} 
We denote the real downstream dataset as $D^{\text{real}}$ and the synthetically generated dataset as $D^{\text{syn}}$. 
Each dataset is partitioned into training, validation, and test subsets, represented as 
$D^{\text{\{real/syn\}}}_{\text{train}}$, 
$D^{\text{\{real/syn\}}}_{\text{val}}$, 
and $D^{\text{\{real/syn\}}}_{\text{test}}$, respectively. 
For context–query splits used during in-context fine-tuning, we write 
$D^{\text{\{real/syn\}}}_{\text{ctx}}$ and 
$D^{\text{\{real/syn\}}}_{\text{query}}$. 
Given that each dataset contains a predefined target column, we represent the non-target feature matrix as 
$X^{\text{\{real/syn\}}}_{\text{train}} \in \mathbb{R}^{n \times d}$ 
and the corresponding target vector as 
$Y^{\text{\{real/syn\}}}_{\text{train}} \in \mathbb{R}^{n \times 1}$. As there is a optional normalization step, between the training samples and the generator data, we sometimes write $D_\text{generator}$, to specify this. 
This notation is used consistently throughout the methodology and experimental sections.

\section{Generator Details}
\label{app:generator_details}
In this section, we describe the different data-generating methods, which are used for the experiments. We start with our baseline, which uses $D_{train}^{real}$ directly, followed by a heuristic method and then go into the methods, optimized on the dataset. 

\subsection{Default Generator (baseline)}
\begin{figure}[htbp]
  \centering
  \includegraphics[width=0.8\textwidth]{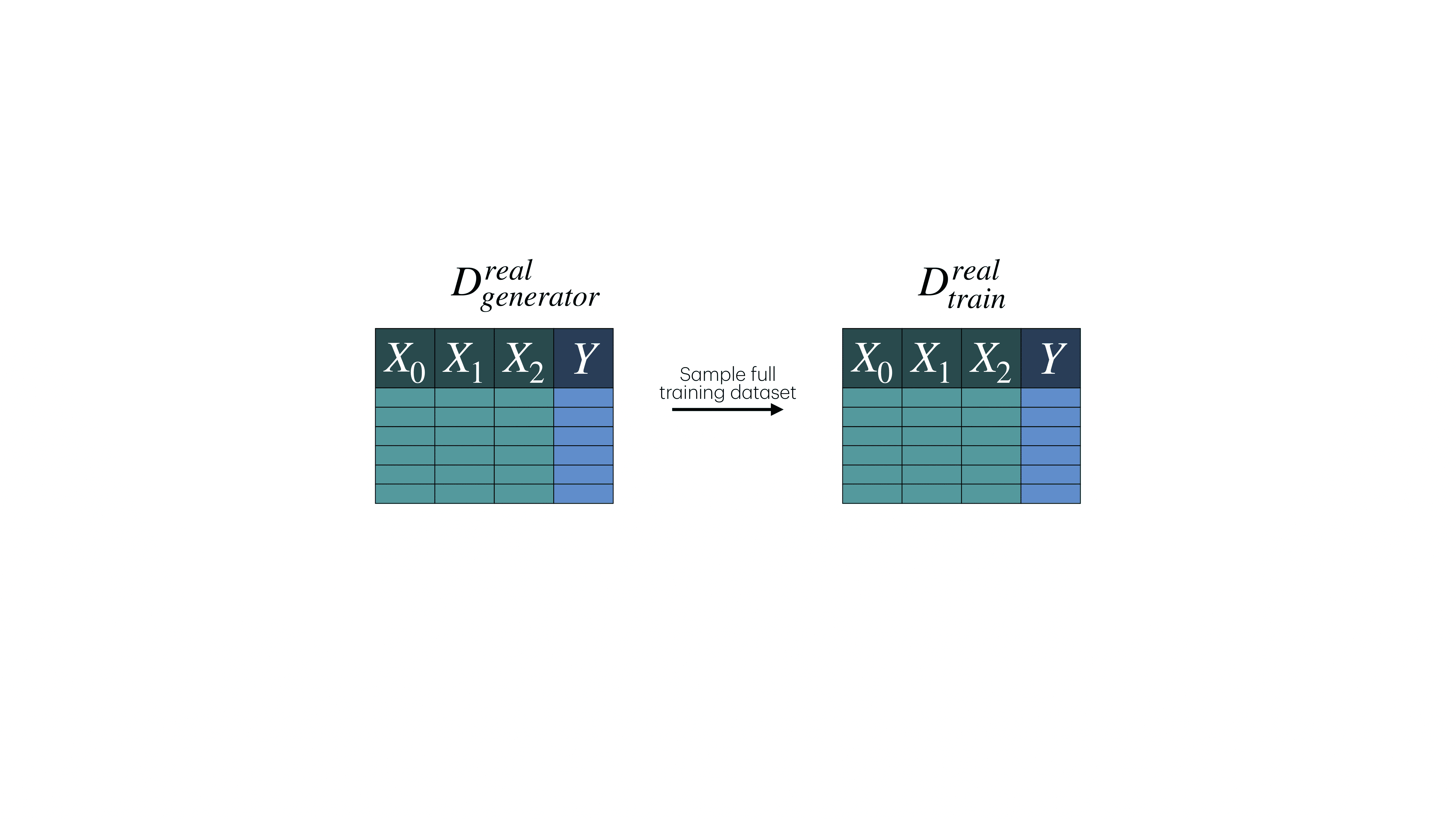}
  \caption{Fine-tuning on training data. The original training dataset is directly utilized for fine-tuning the model.}
  \label{fig:default_generator}
\end{figure}

In our experiments, fine-tuning directly on the raw training data without any form of augmentation represents our baseline. This approach reflects the standard practice of utilizing available data for fine-tuning, as employed in prior work \cite{breejen2024tabforest, den2023finetuning_retrieval, rubachev2025on_finetuning_tabFM}. Since the experiments focus on tiny-to-small datasets, all foundation models considered are capable of processing the entire dataset within one forward pass, without requiring any context retrieval mechanisms.

\subsection{TableAugment Generator}
\begin{figure}[htbp]
  \centering
  \includegraphics[width=0.8\textwidth]{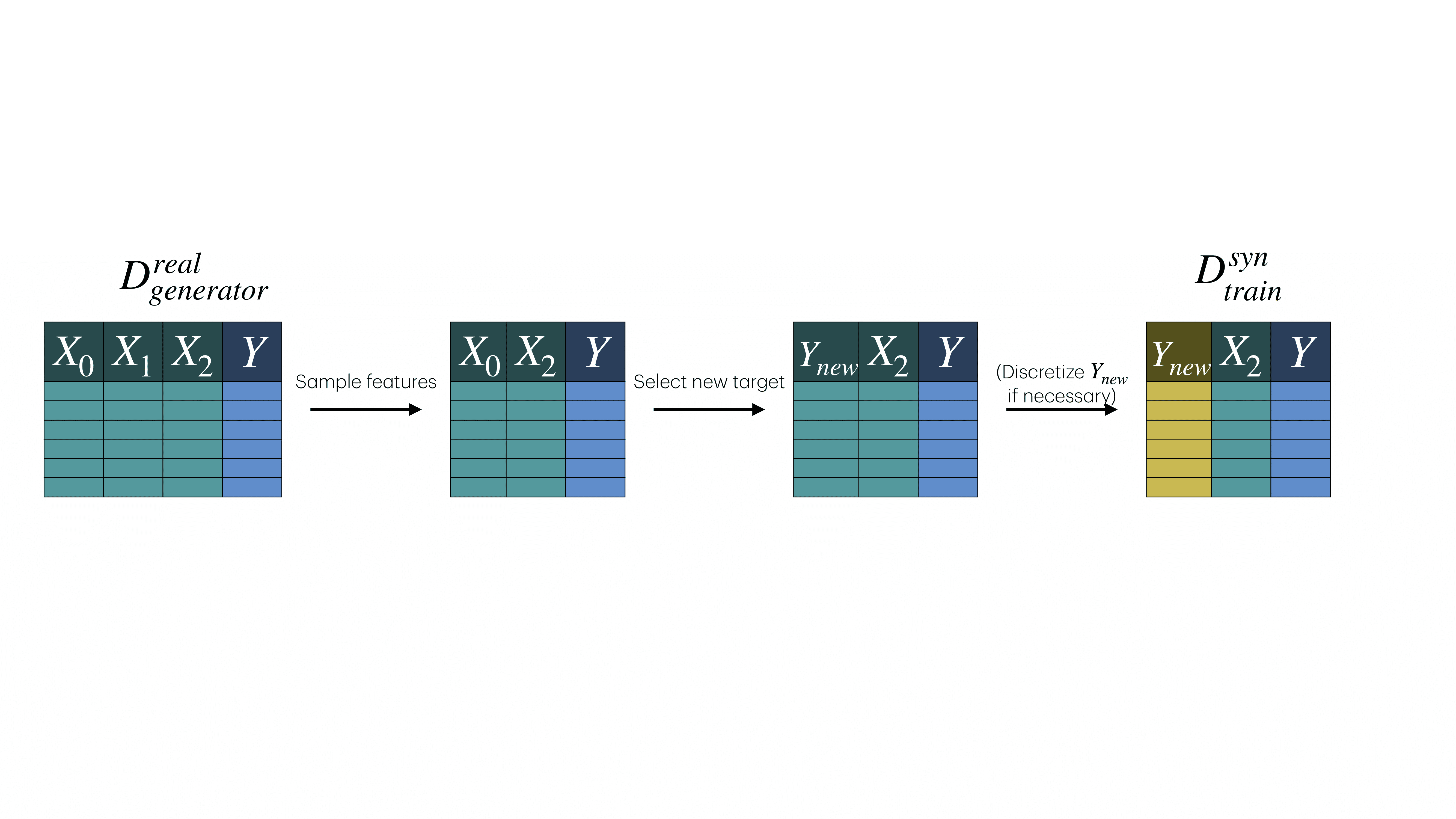}
  \caption{The TableAugment Generator. Real data features are green and blue, while synthetized features are orange and yellow.}
  \label{fig:table_augmentation_generator}
\end{figure}

The TableAugment generator is inspired by the data augmentation approach introduced by \citet{ma2024tabdpt}, who used real-world tabular datasets to pre-train a TabPFN variant. Due to the limited number of publicly available tabular datasets, the authors augmented their collection using different views of each dataset by subsampling and shuffling features and random selection of a target column in each iteration.

In contrast to their pre-training method, our goal is to improve the model's performance on one specific target dataset, and therefore we only need to augment the same dataset over and over again. To this end, our implementation supports a range of configurable augmentation strategies involving feature subselection and target column assignment.

\textbf{Feature Subseletion.} \quad
Feature subselection can be toggled on or off. If disabled, the model uses all features from the generator dataset $D^{real}_{\text{generator}}$. If enabled, a subset of features is selected in each iteration by uniformly sampling a proportion of features between 50\% and 100\%. Additionally, we provide control over the inclusion of the original target column (hereafter referred to as the “old target”) within the selected feature subset. The old target can be configured to be always included, never included, or included with the same probability as all other features.

\textbf{Target Column Selection.} \quad
After the feature subselection, we assign a new target column for the foundation model to predict in this iteration. This functionality can also be enabled or disabled. If disabled, the old target remains the target throughout all iterations. If enabled, a new target column is sampled randomly from the set of selected features. The inclusion of the old target column in the candidate pool for new targets is controllable: it can always be included, never included, or included at random. If the newly selected target column is continuous or exhibits high cardinality, it may need to be discretized. The number of discretized classes, denoted by $\hat{c}$, can either match the number of classes in the old target or be sampled uniformly from a user-specified range (default range is between 2 and 10). To discretize a target column into $\hat{c}$ classes, we assign the $\hat{c} - 1$ most frequent values to individual classes and group all remaining values into the $\hat{c}^{\text{th}}$ class.

The new resulting dataset $D^{syn}$ is the composed of the subselected features and the new selected target column. 

\subsection{Mixed-Model Generator}

\begin{figure}[htbp]
  \centering
  \includegraphics[width=0.8\textwidth]{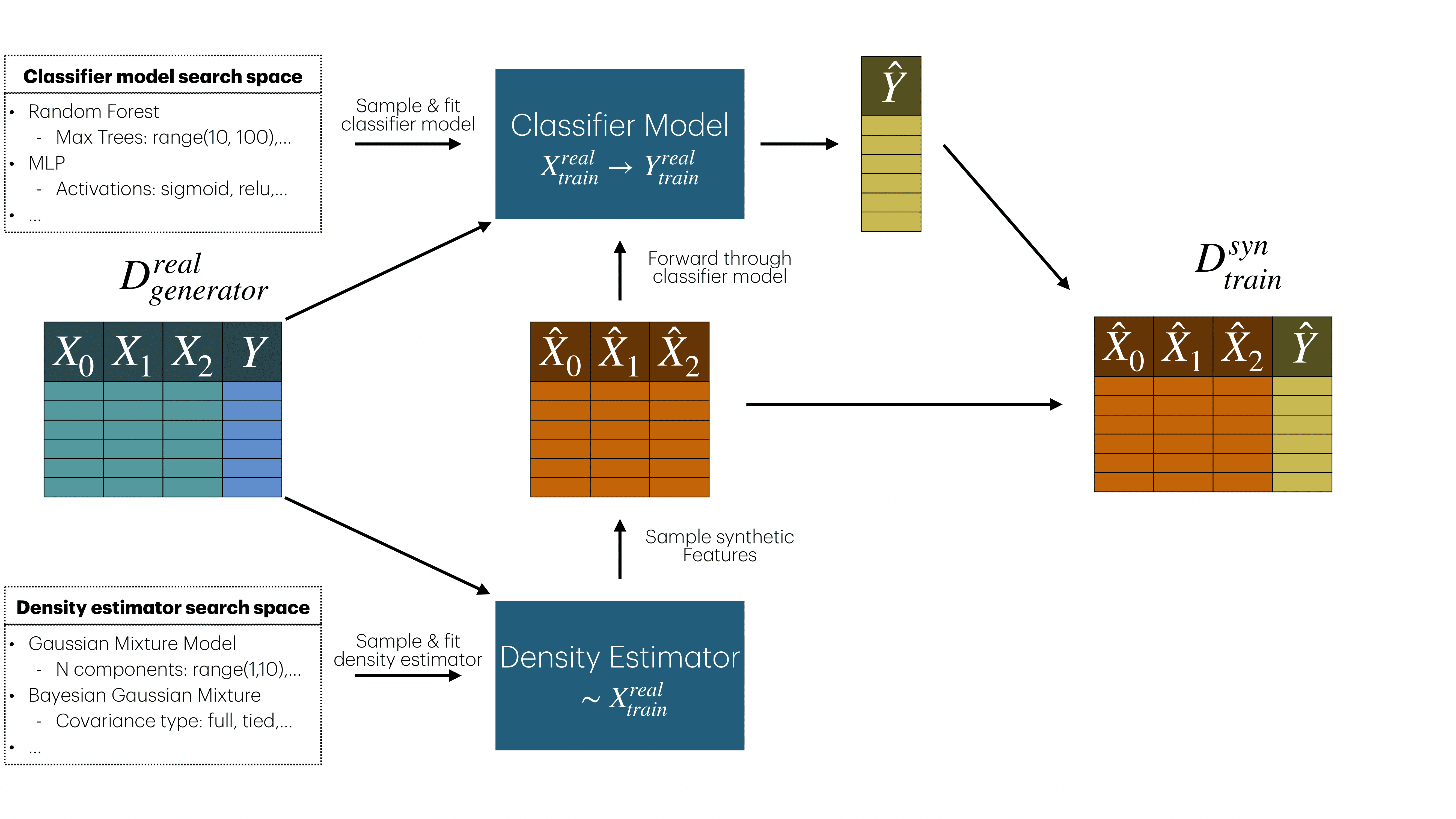}
  \caption{The Mixed-Model generator. First the internal density estimator and classifier model are sampled from predefined search spaces and fitted on $D_{generator}^{real}$. Secondly, to generate synthetic data, we sample features from the density estimator and propagate the samples through the classifier, which yields the labels. The features and labels are concatenated, which results in the synthetic dataset.}
  \label{fig:mixed_model_generator}
\end{figure}

The Mixed-Model Generator is the first proposed augmentation method that incorporates learnable internal models to generate synthetic datasets, as shown in figure \ref{fig:mixed_model_generator}. \citet{breejen2024tabforest} showed that during pre-training data generated through machine learning models can capture complex feature-target relations and thus is very efficient for improving in-context learning capabilities. Based on this insight, we create a generator, which leverages internal machine learning models to generate synthetic datasets, which incorporate information about the real feature to target mapping. It consists of two primary components:
\begin{itemize}
    \item A density estimator, which models the distribution of the feature space.
    \item A classifier, which learns mapping between the features and targets.
\end{itemize}

Together, these components are used to produce labeled synthetic datasets. The generator exposes a range of hyperparameters that control the behavior of both components, as described below.

\textbf{Density Estimator.} \quad
We support four types of density estimators: Gaussian Mixture Model (GMM), Bayesian Gaussian Mixture (BGM), Kernel Density Estimation (KDE), and Uniform Density Model.
The first three options use the implementations provided by \textit{scikit-learn}, while the uniform estimator is custom implemented. For the uniform estimator, continuous features are sampled from a uniform float distribution bounded by the observed range in the training data, while categorical features are sampled uniformly from the set of observed integer values.

A special case is handled for the Bayesian Gaussian Mixture model. When the covariance matrix becomes singular (which can occur in highly imbalanced datasets), we iteratively increase the \textit{cov\_reg} parameter by a factor of 10, up to 10 times. If this fails to resolve the issue, we default to using the uniform density estimator.

\textbf{Classifier Model.} \quad
Once the density estimator is selected and fitted, we sample a classifier model from the following set: Decision Tree (DT), Random Forest (RF), Gradient Boosted Trees (GradBoost), Support Vector Classifier (SVC) and Multi-Layer Perceptron (MLP). Each classifier is associated with a predefined set of hyperparameters, which are shown in (TODO).

\textbf{Synthetic Data Generation.} \quad
To generate synthetic data, we first sample a density estimator with corresponding hyperparameters and a classifier with corresponding hyperparameters. The density estimator is trained on the generator's real feature set, $X_{\text{generator}}^{\text{real}}$, while the classifier is trained using both the real features $X_{\text{generator}}^{\text{real}}$ and the corresponding real targets $Y_{\text{generator}}^{\text{real}}$. Further, we sample $\hat{n}_i$ new feature samples from the density estimator (default: 20,000). These features are forwarded through the trained classifier to produce the corresponding synthetic labels $Y^{\text{syn}}$, forming a complete synthetic dataset $(X^{\text{syn}}_{train}, Y^{\text{syn}}_{train})$.

\subsection{SCM-Based Generator}
\begin{figure}[htbp]
  \centering
  \includegraphics[width=0.8\textwidth]{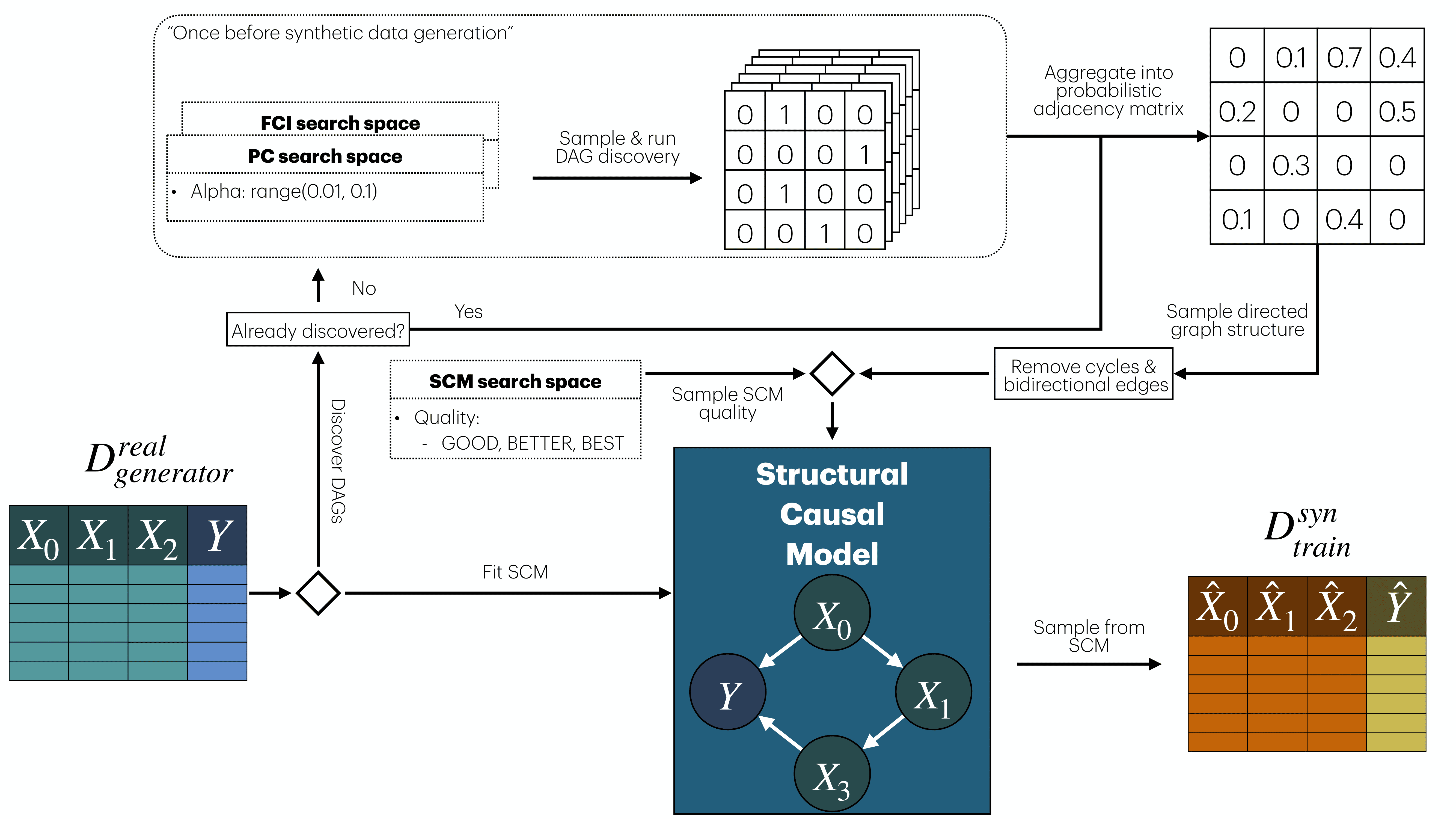}
  \caption{The SCM generator consists of two phases. First, discovering the structural relationships between the features, and secondly sampling DAGs and fitting SCMs on the generator dataset $D_{generator}^{real}$.}
  \label{fig:scm_generator}
\end{figure}
In this method, we estimate dependencies between features through structure discovery and use Structural Causal Models (SCMs) to generate synthetic datasets. The process consists of two phases: first we discover the structural relationships between the features, and secondly we sample and fit SCMs from which we sample synthetic data.

\textbf{Structural Dependencies Discovery.} \quad
In the first phase, which we only have to do once per fine-tuning run, we want to find structural dependencies between the features. Therefore, we apply the Peter-Clark (PC) and Fast Causal Inference (FCI) algorithms from the \texttt{causal-learn} library, each executed 50 times with differently sampled hyperparameters, resulting in a total of 100 discovery runs. The procedure is as follows:

\begin{itemize}
    \item Each run is limited to a maximum runtime of 20 minutes to avoid infinite loops or long convergence times.
    \item The input data is subsampled to a maximum of 1,000 rows and 50 columns if these thresholds are exceeded.
    \item Each run returns an adjacency matrix that indicates detected edges between features.
    \item These 100 adjacency matrices are aggregated into a probabilistic adjacency matrix $C$, where each cell $c_{i,j}$ denotes the relative frequency with which an edge from feature $i$ to feature $j$ was discovered:
    \[
    c_{i,j} = \frac{\text{Number of runs where edge } i \rightarrow j \text{ was found}}{\text{Total number of runs}}.
    \]
\end{itemize}

Although PC and FCI are typically used to recover causal graphs under strict assumptions, our use case only requires discovery of correlational structure. As such, potential violations of assumptions are acceptable, and the discovered graphs are treated as representations of meaningful (though not necessarily causal) dependencies. This step is performed once during initialization. Section~\ref{fig:probabilistic_adjacency} shows an exemplary probabilistic adjacency matrix. 

\textbf{SCM Fitting and Data Generation.}\quad Once the probabilistic adjacency matrix is computed, the following steps are performed each time a synthetic dataset is generated:

\begin{itemize}
    \item \textbf{Graph Sampling:}
    \begin{itemize}
        \item A directed graph is sampled from the probabilistic adjacency matrix $C$.
        \item Bidirectional edges are resolved by randomly removing one direction.
        \item Cycles are removed by randomly deleting edges until the graph becomes a Directed Acyclic Graph (DAG).
    \end{itemize}

    \item \textbf{SCM Fitting:}
    \begin{itemize}
        \item The resulting DAG is passed to DoWhy’s SCM fitting API, which fits an additive noise model using the structure and the generator data $D_{\text{generator}}^{real}$.
        \item The fitting quality can be configured via a \texttt{quality} parameter, with the following options:
        \begin{itemize}
            \item \textbf{GOOD:} Fast and simple models.
            \begin{itemize}
                \item Numerical: Linear regressors (with/without polynomial features), Histogram Gradient Boost Regressor.
                \item Categorical: Logistic regression (with/without polynomial features), Histogram Gradient Boost Classifier.
            \end{itemize}
            \item \textbf{BETTER:} Wider model variety for better accuracy.
            \begin{itemize}
                \item Numerical: Adds Ridge, Lasso, Random Forest, SVR, Extra Trees, KNN, AdaBoost.
                \item Categorical: Adds Random Forest, Extra Trees, SVC, KNN, Gaussian Naive Bayes, AdaBoost.
            \end{itemize}
            \item \textbf{BEST:} Uses \texttt{AutoGluon} (AutoML). Offers highest accuracy but slower training and inference.
        \end{itemize}
    \end{itemize}
    \end{itemize}
Synthetic samples are generated using DoWhy’s API by drawing noise for exogenous variables and propagating it through the SCM. The default number of samples per synthetic dataset is set to 20,000.

\subsection{TabEBM Generator}
The TabEBM generator is a class conditional generating method. Using the TabEBM official implementation, this returns data, where each class is present with the same number of samples. To have a dataset, which is more representative of the real training data, we subsample from the synthetic dataset, such that the class distribution of the real training dataset is maintained. After this, the subsampled dataset is returned. 
As the sampling of TabEBM works purely through in-context learning, the TabEBM generator needs to be fitted every time we generate a new synthetic dataset. 
\begin{figure}[htbp]
  \centering
  \includegraphics[width=0.8\textwidth]{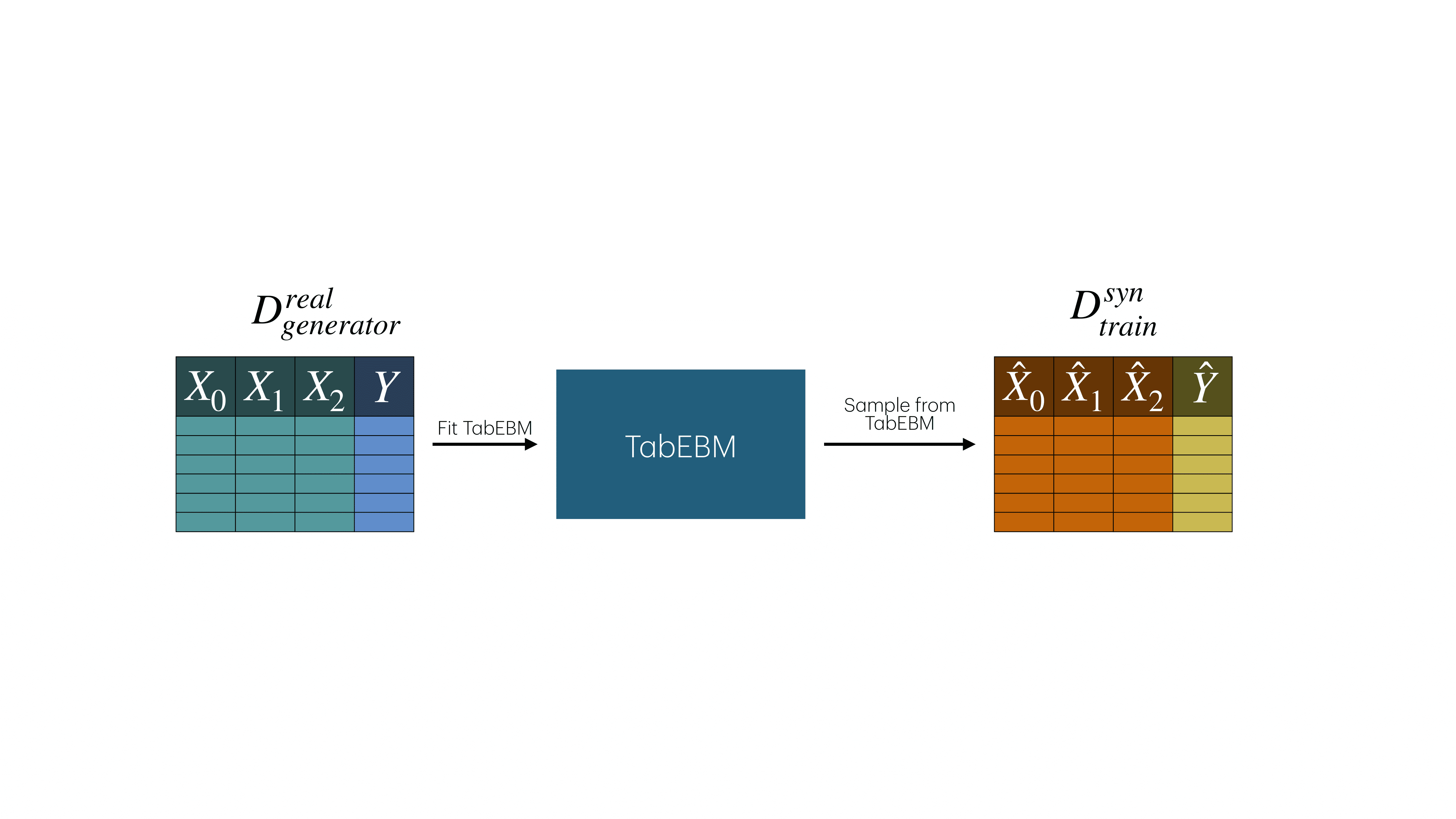}
  \caption{The TabEBM generator.}
  \label{fig:tabebm_generator}
\end{figure}

\subsection{CTGAN Generator}
\begin{figure}[htbp]
  \centering
  \includegraphics[width=0.8\textwidth]{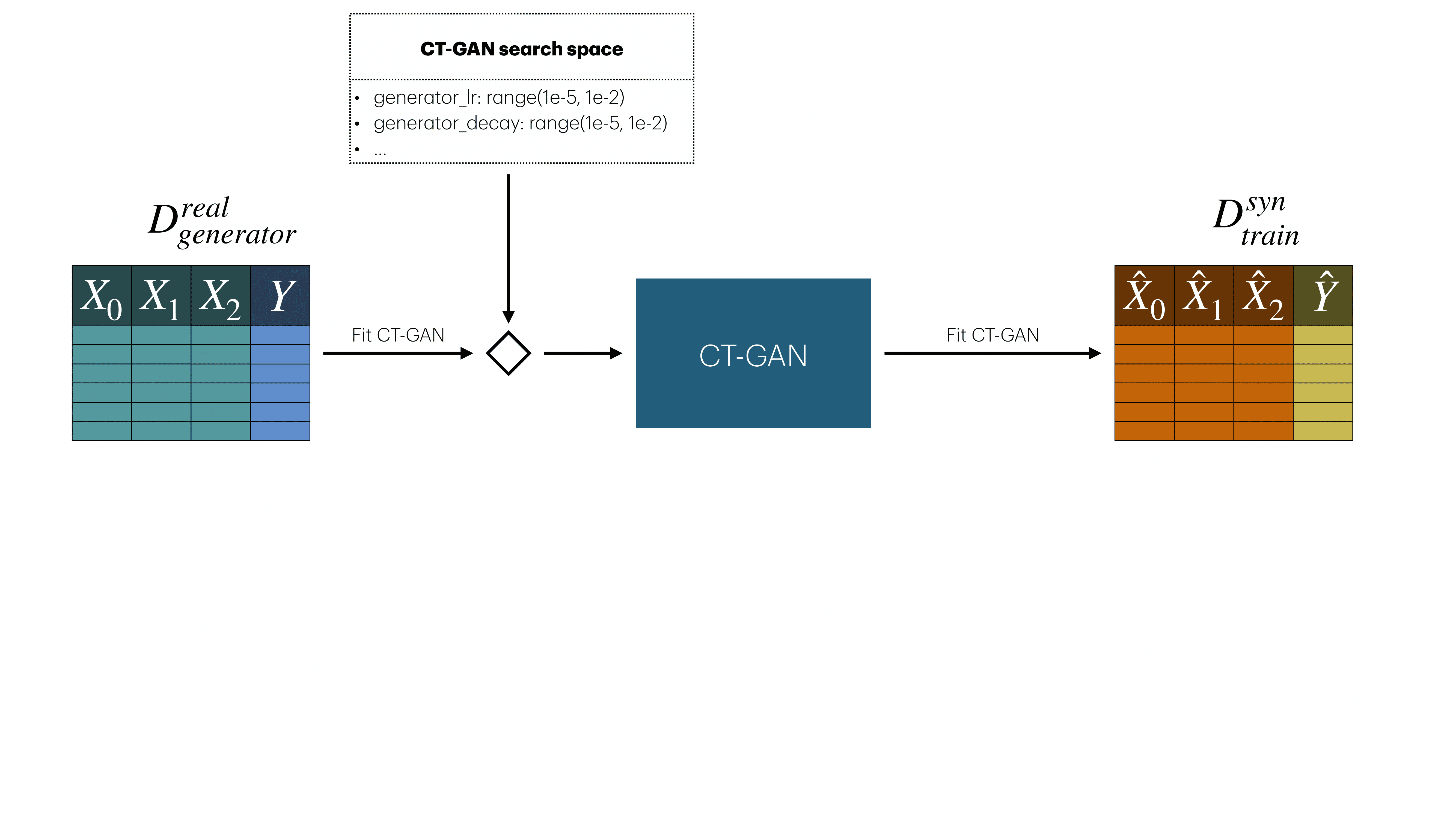}
  \caption{The CT-GAN generator.}
  \label{fig:ctgan_generator}
\end{figure}
For the CTGAN generator, before generating any synthetic data, we fit the GAN on the training data using a set of sampled hyperparameters from a predefined search space. Once the GAN model is fitted, we can directly sample synthetic data from it (default 20,000 samples), which represent the synthetic dataset.

\section{Generator Hyperparameters}
\subsection{MixedModel Generator Hyperparameters} 
\begin{table}[H]
    \centering
    \caption{Overview of hyperparameter ranges for classifiers used in the \textit{MixedModel generator}.}
    \label{tab:hyperparameter_ranges}
    \resizebox{\textwidth}{!}{
    \begin{tabular}{@{}llll@{}}
        \toprule
        \textbf{Classifier} & \textbf{Hyperparameter} & \textbf{Type / Choices} & \textbf{Range or Values} \\
        \midrule
        \multirow{3}{*}{TabPFNClassifier} 
            & \texttt{n\_estimators} & Integer (Uniform) & [1, 10] \\
            & \texttt{n\_jobs} & Categorical & \{1\} \\
            & \texttt{device} & Categorical & \{"cpu"\} \\
        \midrule
        \multirow{7}{*}{RandomForestClassifier} 
            & \texttt{n\_estimators} & Integer (Log-Uniform) & [10, 500] \\
            & \texttt{criterion} & Categorical & \{"gini", "log\_loss", "entropy"\} \\
            & \texttt{max\_depth} & Integer (Log-Uniform) & [10, 100] \\
            & \texttt{min\_samples\_split} & Integer (Uniform) & [2, 20] \\
            & \texttt{min\_samples\_leaf} & Integer (Uniform) & [1, 10] \\
            & \texttt{max\_leaf\_nodes} & Integer (Uniform) & [10, 100] \\
            & \texttt{bootstrap} & Categorical & \{True, False\} \\
        \midrule
        \multirow{6}{*}{DecisionTreeClassifier}
            & \texttt{criterion} & Categorical & \{"gini", "entropy", "log\_loss"\} \\
            & \texttt{splitter} & Categorical & \{"best", "random"\} \\
            & \texttt{max\_depth} & Integer (Log-Uniform) & [5, 100] \\
            & \texttt{min\_samples\_split} & Integer (Uniform) & [2, 20] \\
            & \texttt{min\_samples\_leaf} & Integer (Uniform) & [1, 10] \\
            & \texttt{max\_features} & Categorical & \{0.1, 0.25, 0.5, 0.75, 1.0, "sqrt", "log2", None\} \\
        \midrule
        \multirow{10}{*}{MLPClassifier}
            & \texttt{hidden\_layer\_sizes} & Integer (Uniform) & [1, 100] \\
            & \texttt{activation} & Categorical & \{"relu", "logistic", "tanh"\} \\
            & \texttt{solver} & Categorical & \{"adam", "sgd", "lbfgs"\} \\
            & \texttt{alpha} & Float (Uniform) & [0.0001, 0.1] \\
            & \texttt{batch\_size} & Categorical & \{"auto", 32, 64, 128\} \\
            & \texttt{learning\_rate} & Categorical & \{"constant", "invscaling", "adaptive"\} \\
            & \texttt{learning\_rate\_init} & Float (Uniform) & [0.0001, 0.01] \\
            & \texttt{max\_iter} & Integer (Uniform) & [100, 1000] \\
            & \texttt{momentum} & Float (Uniform) & [0.5, 0.95] \\
            & \texttt{nesterovs\_momentum / early\_stopping} & Categorical & \{True, False\} \\
        \midrule
        \multirow{11}{*}{SVC}
            & \texttt{kernel} & Categorical & \{"linear", "rbf", "poly", "sigmoid"\} \\
            & \texttt{C} & Float (Log-Uniform) & [1e-6, 1e6] \\
            & \texttt{degree} & Integer (Uniform) & [1, 5] \\
            & \texttt{gamma} & Categorical & \{"scale", "auto"\} \\
            & \texttt{coef0} & Float (Uniform) & [-1, 1] \\
            & \texttt{shrinking} & Categorical & \{True, False\} \\
            & \texttt{probability} & Categorical & \{True, False\} \\
            & \texttt{tol} & Float (Log-Uniform) & [1e-5, 1e-2] \\
            & \texttt{cache\_size} & Float (Uniform) & [200, 1000] \\
            & \texttt{class\_weight} & Categorical & \{None, "balanced"\} \\
            & \texttt{max\_iter / break\_ties} & Integer / Bool & [100, 1000] / \{True, False\} \\
        \midrule
        \multirow{8}{*}{HistGradientBoostingClassifier}
            & \texttt{loss} & Categorical & \{"log\_loss"\} \\
            & \texttt{learning\_rate} & Float (Uniform) & [0.01, 1.0] \\
            & \texttt{max\_iter} & Integer (Uniform) & [50, 1000] \\
            & \texttt{max\_leaf\_nodes} & Integer (Uniform) & [5, 100] \\
            & \texttt{max\_depth} & Integer (Uniform) & [3, 15] \\
            & \texttt{min\_samples\_leaf} & Integer (Uniform) & [5, 100] \\
            & \texttt{l2\_regularization} & Float (Uniform) & [0.0, 1.0] \\
            & \texttt{max\_bins} & Integer (Uniform) & [10, 255] \\
        \bottomrule
    \end{tabular}
    }
\end{table}

\subsection{SCM Generator Hyperparameters}
\begin{table}[H]
    \centering
    \caption{Overview of the \texttt{quality} hyperparameter for internal model assignment in the \textit{SCM generator}.}
    \label{tab:scm_assignment_quality}
    \resizebox{\textwidth}{!}{
    \begin{tabular}{@{}lll@{}}
        \toprule
        \textbf{Quality Setting} & \textbf{Included Models (Examples)} & \textbf{Description / Characteristics} \\
        \midrule
        \texttt{GOOD} &
        \begin{tabular}[c]{@{}l@{}}
            Numerical: Linear Regressor, Polynomial Regressor, \\ 
            Histogram Gradient Boost Regressor \\
            Categorical: Logistic Regressor, Polynomial Logistic Regressor, \\ 
            Histogram Gradient Boost Classifier
        \end{tabular} &
        Small, efficient model set for fast training and inference; 
        medium predictive accuracy. \\
        \midrule
        \texttt{BETTER} &
        \begin{tabular}[c]{@{}l@{}}
            Numerical: Ridge, Lasso, Random Forest, SVR, \\ 
            Extra Trees, KNN, AdaBoost \\
            Categorical: Random Forest, Extra Trees, SVC, \\ 
            KNN, GaussianNB, AdaBoost
        \end{tabular} &
        Expanded model pool for higher accuracy while maintaining reasonable training speed. \\
        \midrule
        \texttt{BEST} &
        \begin{tabular}[c]{@{}l@{}}
            AutoML backend (AutoGluon)
        \end{tabular} &
        Full AutoML configuration offering the best accuracy, but with increased computational cost. \\
        \bottomrule
    \end{tabular}
    }
\end{table}

\subsection{TabEBM Generator Hyperparameters}
\begin{table}[H]
    \centering
    \caption{Overview of hyperparameter ranges for the \textit{TabEBMGenerator}.}
    \label{tab:tabebm_hyperparameter_ranges}
    \resizebox{\textwidth}{!}{
    \begin{tabular}{@{}llll@{}}
        \toprule
        \textbf{Generator} & \textbf{Hyperparameter} & \textbf{Type / Choices} & \textbf{Range or Values} \\
        \midrule
        \multirow{3}{*}{TabEBMGenerator} 
            & \texttt{n\_samples\_per\_class} & Integer (Fixed) & 150 \\
            & \texttt{device} & Categorical & \{"cpu"\} \\
            & \texttt{name} & Categorical & \{"TabEBMGenerator"\} \\
        \bottomrule
    \end{tabular}
    }
\end{table}
\subsection{CTGAN Generator Hyperparameters}
    \begin{table}[H]
        \centering
        \caption{Overview of hyperparameter ranges for the \textit{CTGANGenerator}.}
        \label{tab:ctgan_hyperparameter_ranges}
        \resizebox{\textwidth}{!}{
        \begin{tabular}{@{}llll@{}}
            \toprule
            \textbf{Generator} & \textbf{Hyperparameter} & \textbf{Type / Choices} & \textbf{Range or Values} \\
            \midrule
            \multirow{13}{*}{CTGANGenerator}
                & \texttt{refit\_interval} & Integer (Fixed) & 10 \\
                & \texttt{n\_synthetic\_samples} & Integer (Fixed) & 20{,}000 \\
                & \texttt{n\_sample\_attempts} & Integer (Fixed) & 10 \\
                & \texttt{model\_cache\_lower\_bound} & Integer (Fixed) & 2 \\
                & \texttt{model\_cache\_upper\_bound} & Integer (Fixed) & 5 \\
                & \texttt{cuda} & Categorical & \{True\} \\
                & \texttt{embedding\_dim} & Integer (Uniform) & [8, 256] \\
                & \texttt{generator\_lr} & Float (Log-Uniform) & [1e-5, 1e-2] \\
                & \texttt{generator\_decay} & Float (Log-Uniform) & [1e-5, 1e-2] \\
                & \texttt{discriminator\_lr} & Float (Log-Uniform) & [1e-5, 1e-2] \\
                & \texttt{discriminator\_decay} & Float (Log-Uniform) & [1e-5, 1e-2] \\
                & \texttt{discriminator\_steps} & Integer (Uniform) & [1, 10] \\
                & \texttt{epochs} & Integer (Uniform) & [100, 200] \\
            \bottomrule
        \end{tabular}
        }
    \end{table}

\subsection{TableAugment Generator Hyperparameters}
    \begin{table}[H]
        \centering
        \caption{Overview of hyperparameter ranges for the \textit{TableAugmentGenerator}.}
        \label{tab:TableAugment_hyperparameter_ranges}
        \resizebox{\textwidth}{!}{
        \begin{tabular}{@{}llll@{}}
            \toprule
            \textbf{Generator} & \textbf{Hyperparameter} & \textbf{Type / Choices} & \textbf{Range or Values} \\
            \midrule
            \multirow{10}{*}{TableAugmentGenerator}
                & \texttt{name} & Categorical & \{"TableAugmentGenerator"\} \\
                & \texttt{normalize} & Categorical & \{False\} \\
                & \texttt{sub\_sample\_features.active} & Categorical & \{True\} \\
                & \texttt{sub\_sample\_features.min\_ratio} & Float (Uniform) & [0.5, 1.0] \\
                & \texttt{sub\_sample\_features.max\_ratio} & Float (Uniform) & [0.5, 1.0] \\
                & \texttt{sub\_sample\_features.include\_target} & Categorical & \{"random", "always", "never"\} \\
                & \texttt{random\_sample\_target.active} & Categorical & \{True\} \\
                & \texttt{random\_sample\_target.include\_target} & Categorical & \{"random", "always", "never"\} \\
                & \texttt{random\_sample\_target.allow\_target\_as\_target} & Categorical & \{True\} \\
                & \texttt{random\_sample\_target.use\_dataset\_num\_classes} & Categorical & \{True\} \\
                & \texttt{random\_sample\_target.min\_discrete\_values} & Integer (Fixed) & 2 \\
                & \texttt{random\_sample\_target.max\_discrete\_values} & Integer (Fixed) & 10 \\
            \bottomrule
        \end{tabular}
        }
    \end{table}

\section{Data-Splitting}
\label{sec:data_splitting}
\begin{figure}[htbp]
  \centering
  \includegraphics[width=0.8\textwidth]{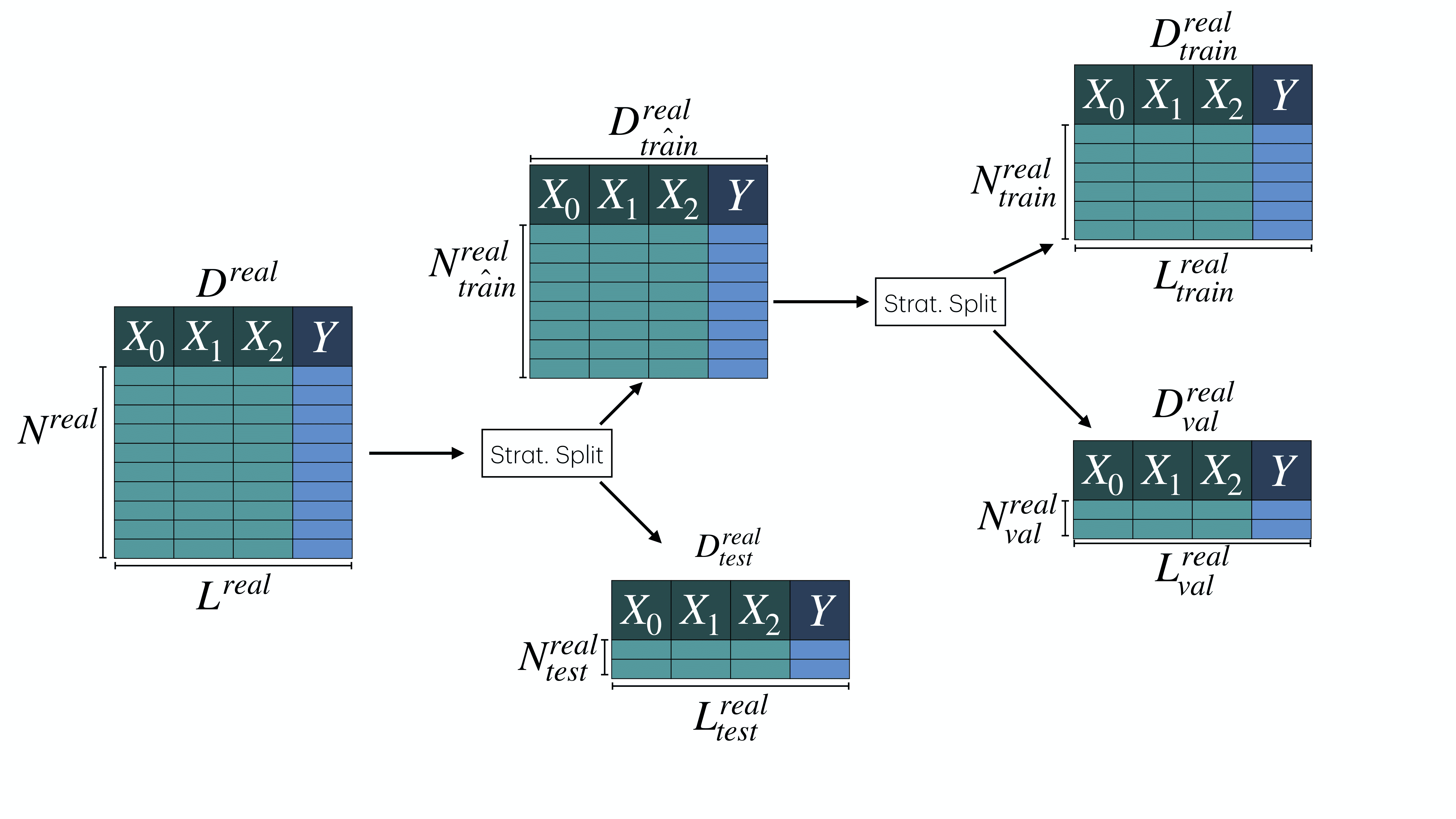}
  \caption{Splitting the full dataset into a train, validation and test set.}
  \label{fig:data_splitting}
\end{figure}
Given a target dataset \( D^{\text{real}} \), we partition it into three mutually exclusive subsets: training, validation, and test. Formally, this decomposition is expressed as:
\[
D^{\text{real}} = D^{\text{real}}_{\text{train}} \cup D^{\text{real}}_{\text{val}} \cup D^{\text{real}}_{\text{test}}, \quad \text{with} \quad D^{\text{real}}_{\text{train}} \cap D^{\text{real}}_{\text{val}} = \emptyset, \quad D^{\text{real}}_{\text{train}} \cap D^{\text{real}}_{\text{test}} = \emptyset, \quad D^{\text{real}}_{\text{val}} \cap D^{\text{real}}_{\text{test}} = \emptyset.
\]

To evaluate the performance of fine-tuning under limited data conditions, we constrain the dataset sizes used for training and validation. Including, if the total number of samples \( N^{\text{real}} \) exceeds 1000, we apply truncation as follows:

\begin{itemize}
    \item The training set \( D^{\text{real}}_{\text{train}} \in \mathbb{R}^{N^{\text{real}}_{\text{train}} \times L^{\text{real}}} \), with
    \[
    N^{\text{real}}_{\text{train}} = \min(0.6 \cdot N^{\text{real}}, 600).
    \]
    
    \item The validation set \( D^{\text{real}}_{\text{val}} \in \mathbb{R}^{N^{\text{real}}_{\text{val}} \times L^{\text{real}}} \), with
    \[
    N^{\text{real}}_{\text{val}} = \min(0.2 \cdot N^{\text{real}}, 200).
    \]
    
    \item The test set \( D^{\text{real}}_{\text{test}} \in \mathbb{R}^{N^{\text{real}}_{\text{test}} \times L^{\text{real}}} \), where
    \[
    N^{\text{real}}_{\text{test}} = \max(0.2 \cdot N^{\text{real}}, N^{\text{real}} - N^{\text{real}}_{\text{train}} - N^{\text{real}}_{\text{val}}).
    \]
\end{itemize}
Regarding the validation split, we first apply a test, $\hat{\text{train}}$ split, and then get the train, val split only based on $\hat{\text{train}}$.
This procedure ensures that the training and validation subsets comprise at most 60\% and 20\% of the full dataset, capped at 600 and 200 samples respectively. The test set comprises the remaining data, ensuring at least 20\% coverage.

To preserve the distribution of class labels, all splits are generated using stratified sampling with respect to the class labels. This partitioning process is repeated across \( K \) different folds to support robust evaluation. The overall splitting strategy is illustrated in Figure~\ref{fig:data_splitting}.

\section{Data Pre-processing}
 \label{sec:data_preprocessing}

\begin{figure}[htbp]
  \centering
  \includegraphics[width=0.8\textwidth]{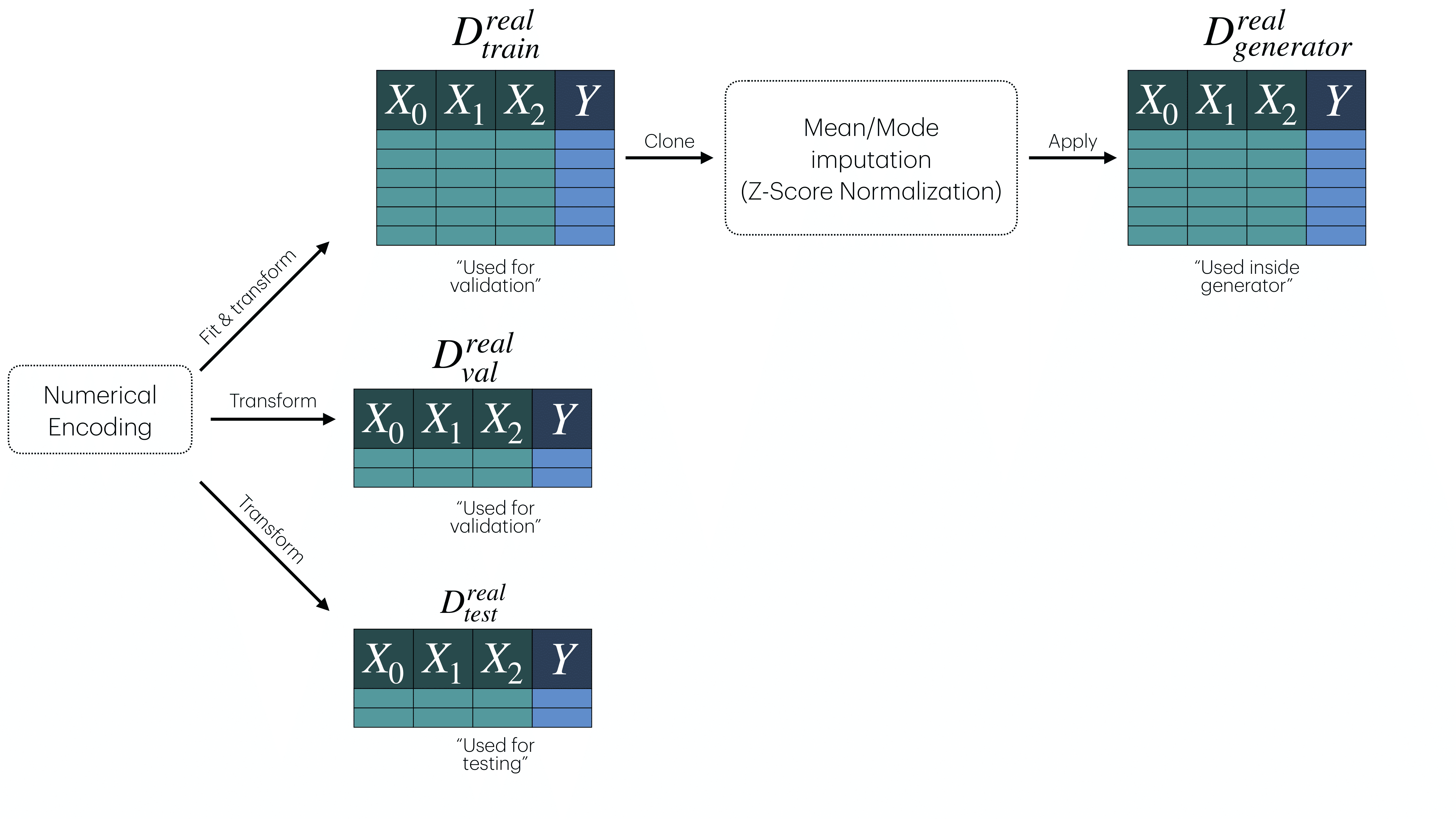}
  \caption{Pre-processing the data. Non-numerical features and numerically encoded. Further, the training data is cloned for the data generating methods with mean/mode imputation and optional z-score normalization applied.}
  \label{fig:data_preprocessing}
\end{figure}

Having obtained the training, validation, and test splits, we proceed with data pre-processing to ensure compatibility with both the data-generating procedures and the foundation model. 

For the foundation model, all non-numerical (categorical or textual) features are encoded into numerical representations to enable model compatibility. This step is essential for ensuring that the input conforms to the expected numerical format of the foundation model. For this, we use Autogluon's \textit{AutoMLPipelineFeatureGenerator}, with the default setting. 

Further, for the data-generating methods, we create a working copy of the training dataset $D^{\text{real}}_{\text{train}} \rightarrow D^{\text{real}}_{\text{generator}}$. On this cloned dataset, we perform the following pre-processing steps:

\begin{itemize}
    \item \textbf{Mean imputation} is applied to continuous (real-valued) features to handle missing values.
    \item \textbf{Mode imputation} is used for categorical features, ensuring that missing entries are filled with the most frequent category.
    \item Optionally, \textbf{Z-score normalization} is applied to standardize the input distribution.
\end{itemize}
During the fine-tuning we use $D^{\text{real}}_{\text{generator}}$ to build our generators, and $D^{\text{real}}_{\text{train}}, D^{\text{real}}_{\text{val}}$ to assess validation performance. 
This pre-processing workflow is illustrated in Figure~\ref{fig:data_preprocessing}.

\section{Fine-tuning Hyperparameters} 
    \label{subsec:finetuning_details}
    \begin{table}[H]
        \centering
        \caption{Default parameter values of the training hyperparameter configuration.}
        \begin{tabular}{@{}lc@{}}
            \toprule
            \textbf{Parameter} & \textbf{Default Value} \\ 
            \midrule
            \texttt{initial\_learning\_rate} & $1\times10^{-4}$ \\
            \texttt{finetune\_steps} & 50 \\
            \texttt{shuffle\_classes} & False \\
            \texttt{shuffle\_features} & False \\
            \texttt{use\_random\_transforms} & False \\
            \texttt{random\_mirror\_x} & True \\
            \texttt{patience} & 40 \\
            \bottomrule
        \end{tabular}
        \label{tab:finetuning_hps}
    \end{table}

\section{Probabilistic Adjacency Matrix}

In the first step of our method, we apply a set of causal discovery algorithms with varying hyperparameters and aggregate their outputs into a probabilistic adjacency matrix. Each matrix entry represents the relative frequency of a directed edge across all discovered graphs.

\begin{figure}[t]
\centering
\includegraphics[width=0.96\textwidth]{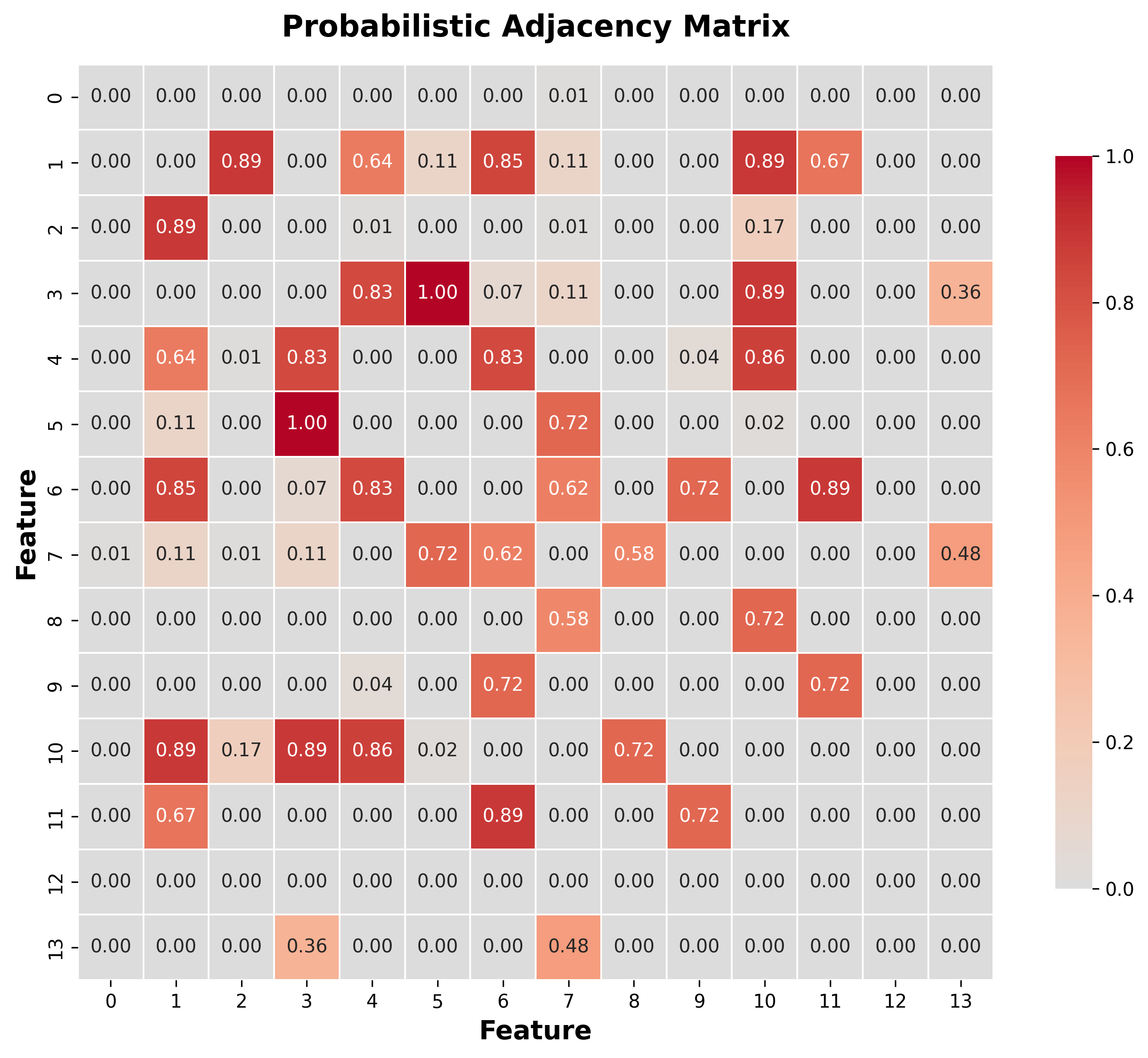}
\caption{Example probabilistic adjacency matrix for the *Is-this-a-good-customer* dataset (fold 5). Each entry encodes the empirical frequency of a directed edge across multiple causal discovery runs with varying hyperparameters. Brighter values indicate higher consensus about the presence and direction of an edge. The overall sparsity reflects the typically sparse causal structure of real-world tabular data.}

\label{fig:probabilistic_adjacency}
\end{figure}

Figure~\ref{fig:probabilistic_adjacency} illustrates an example probabilistic adjacency matrix for the Is-this-a-good-customer dataset (fold 5). Qualitatively, we observe that most matrices are relatively sparse across datasets, consistent with prior findings on sparse real-world data structures~\citep{hollmann2022tabpfnv1}. This sparsity facilitates efficient causal structure learning even in high-dimensional settings (up to 200 features in our setup).

\end{document}